\documentclass{article}
\usepackage[cmex10]{amsmath}
\usepackage{graphicx}
\usepackage[UKenglish]{babel}
\usepackage{fullpage}
\usepackage{algorithm}
\usepackage{algorithmic}
\usepackage[caption=false]{subfig}
\usepackage[percent]{overpic}
\usepackage[square, sort, numbers, authoryear]{natbib}
\usepackage{hyperref}

\providecommand{\keywords}[1]{\textbf{\textit{Keywords---}} #1}

\title{Single camera pose estimation using Bayesian filtering and Kinect motion priors}
\author{Michael Burke and Joan Lasenby\\Department of Engineering,\\ University of Cambridge,\\ Cambridge, UK, CB2 1PZ.\footnote{This work was supported by funding from the Council for Scientific and Industrial Research (CSIR), South Africa and the Cambridge Commonwealth Trust under a CSIR-Cambridge Scholarship.}}

\begin{document}

\maketitle




\begin{abstract} Traditional approaches to upper body pose estimation using monocular vision rely on complex body models and a large variety of geometric constraints. We argue that this is not ideal and somewhat inelegant as it results in large processing burdens, and instead attempt to incorporate these constraints through priors obtained directly from training data. A prior distribution covering the probability of a human pose occurring is used to incorporate likely human poses. This distribution is obtained offline, by fitting a Gaussian mixture model to a large dataset of recorded human body poses, tracked using a Kinect sensor. We combine this prior information with a random walk transition model to obtain an upper body model, suitable for use within a recursive Bayesian filtering framework. Our model can be viewed as a mixture of discrete Ornstein-Uhlenbeck processes, in that states behave as random walks, but drift towards a set of typically observed poses. This model is combined with measurements of the human head and hand positions, using recursive Bayesian estimation to incorporate temporal information. Measurements are obtained using face detection and a simple skin colour hand detector, trained using the detected face. The suggested model is designed with analytical tractability in mind and we show that the pose tracking can be Rao-Blackwellised using the mixture Kalman filter, allowing for computational efficiency while still incorporating bio-mechanical properties of the upper body. In addition, the use of the proposed upper body model allows reliable three-dimensional pose estimates to be obtained indirectly for a number of joints that are often difficult to detect using traditional object recognition strategies. Comparisons with Kinect sensor results and the state of the art in 2D pose estimation highlight the efficacy of the proposed approach.\footnote{This technical report is a vastly extended version of preliminary work \citep{Burke14}, containing additional and expanded detail and new experimental results.} 
\end{abstract}

\keywords{Human pose estimation, Mixture Kalman filter, Computer vision, Kinect}

\section{Introduction}

Reliable human pose estimation is a frequently encountered computer vision task, often required for successful vision-based gesture or action recognition systems. Specifically, our goal is to perform gesture recognition for human-robot interaction, which requires the 3D positions of human upper bodies to be tracked. Unfortunately, this is a particularly challenging problem, especially in cluttered environments with potentially moving cameras. 

2D information typically suffices if only static gestures are to be recognised, but 3D information is required for most temporal gesture recognition solutions \citep{Wu99}. Multiple camera motion capture systems can provide 3D measurements with a high level of accuracy, but often require that users wear markers that aid in detection. Stereo camera vision allows for relatively accurate 3D spatial information to be obtained and as a result is commonly used for temporal gesture recognition. This is evidenced by the gesture recognition schemes of \citet{Triesch98}, \citet{Seong-Wang06} and \citet{Nickel07}, which all use stereo vision systems to observe gestures. 3D information can also be obtained using structured light systems such as the Kinect or PrimeSense depth sensor. Unfortunately, while the Xbox Kinect skeleton tracker \cite{xboxtracker} is extremely effective, in many applications, where payloads are limited, this is infeasible, and a body tracking solution relying only on monocular vision would be preferred.

This paper aims to solve the 3D upper body pose estimation problem using images obtained by only a single camera. We propose a novel upper body model, trained using Kinect pose priors and designed with analytical tractability in mind. We show that pose tracking using this model can be Rao-Blackwellised using the mixture Kalman filter, allowing for computational efficiency while still incorporating bio-mechanical properties of the upper body. The model is used within a recursive Bayesian framework to provide reliable estimates of user head, neck, shoulder, elbow and hand locations when only a subset of body joints can be detected.

Face detection is used to determine head position, and provides a skin colour prior that assists in locating hands. Edge-based error correction is proposed to correct potential hand association errors before head and hand measurements are used to estimate upper body pose.

The paper is organised as follows. Section \ref{sec:Background} discusses related work and provides some background to the problem. This is followed by a description of Bayesian filtering for human pose estimation, the introduction of our body models and various tracking algorithms that can be used with these in section \ref{sec:Bayes}. A comparison of the trackers is provided in section \ref{sec:TrackResults}, before we describe how we obtain head and hand measurements from images in section \ref{headhandmeasure}. Results obtained when these measurements are used in conjunction with our model and body tracker are presented in section \ref{Results}, along with a comparison with a recent 2D pose estimation approach \citep{Eichner12}. Finally, conclusions are provided in section \ref{sec:concs}.

\section{Background and related work}
\label{sec:Background}

Effective human pose estimation is required for successful vision-based gesture recognition systems to be deployed. This section describes various approaches to human pose estimation, within the context of gesture recognition.

A vast amount of work has been conducted in the field of human pose estimation using monocular vision. Two approaches to pose estimation from static images have emerged, the first relying on tracking and generative models, and the second on morphological recognition. Morphological recognition techniques can be top-down, where entire bodies are recognised, or bottom-up, where bodies are recognised by locating various body parts or components. \citet{Gavrila96} use a top-down, search-based technique to locate poses by matching contours or edges formed using a generative body model with those in an input image. 

A number of top-down approaches rely on matching extracted silhouettes to a known database. This technique is applied by \citet{Germann2011} who refine matched pose estimates using a set of 3D body part constraints. This approach relies on multiple cameras though, and  the extraction of silhouettes, which can be challenging. Further, the authors note that additional information is required to estimate poses where the arms are close to the body, as silhouettes do not contain sufficient information to do so.

The dominant approach to pose estimation is bottom-up \citep{Yang2011}, using a pictorial structure of body parts with geometric constraints modelling component interactions. \citet{Yang2011} use a family of affinely warped templates and a mixture model capturing contextual relations and produce good pose estimation results at approximately 1 frame a second. A pictorial structure model is also used by \citet{Eichner12}, who detect bodies using a part-based model, segment these bodies using Grabcut \citep{Rother04} and then fit appearance models trained previously using labelled data. This approach also provides good performance, but can be slow, and only works on near frontal and rear viewpoints.

A number of pose estimation techniques use segmentation to locate and extract human bodies. In their work on pose estimation for sign-language videos, \citet{Charles13a} leverage the layering of signers on video to extract bodies using co-segmentation, before estimating joint locations using a selection of random forests trained on a number of previously segmented bodies (labelled using the work of \citet{Buehler09}). Unfortunately, accurate segmentation is slow on general video sequences and not usually feasible for real time applications. 

Bottom-up approaches to pose estimation are also used in tracking-based pose estimation approaches. \citet{Lee04} used a 21 degree-of-freedom generative model of human kinematics, shape and clothing in a data-driven Markov chain Monte Carlo search. Here, visual cues of the face, head and shoulder contours, skin blobs and arm ridges were used to aid importance sampling and drive a Monte Carlo search to feasible 3D pose candidates. Unfortunately, estimating 3D pose estimation from static 2D images results in a number of pose ambiguities as different body configurations can appear similar when viewed from different points. 

Many pose estimation techniques rely on Monte Carlo simulation or particle filtering. Particle filters represent the posterior belief in a state, conditioned on a set of measurements, by a set of random state samples drawn from this distribution. While the particle filter is able to approximate non-Gaussian noise distributions extremely well, it is computationally intensive as motion and observation models need to operate on multiple particles. Moreover, the memory requirements of particle filter algorithms are excessive, as the performance of the algorithm is dependent on the number of particles used. 

In high dimensional state spaces, the effective number of particles required to approximate the posterior belief can become extremely large and the particle filter tends to operate as a traditional optimisation problem when a feasible number of particles is used. In these cases, additional information is often required to constrain the search space and produce good particle estimates.

\citet{Sminchisescu2001} note that many particle filtering algorithms for 3D pose estimation often require the addition of extra noise to assist in the search for minima. They attempt to resolve this by using a complex body model and through careful design of the observation likelihood function, incorporating priors on the anthropometric data of internal proportions, parameter stabilisers, joint limits, and body part penetration avoidance. They also apply covariance scaled sampling to direct the search, which involves combining assumed dynamics with the posterior distribution and growing the prior covariances to sample more broadly. This search can be sped up through the addition of kinematic reasoning to assist in the sampling, reducing the number of possible solutions to a pose if the lengths of limbs are known \citep{Sminchisescu03}. 

\citet{Jauregui2010} also apply kinematic reasoning to aid in pose estimation, but use a silhouette-based observation model. Here, silhouettes are extracted using background subtraction, faces detected and a skin colour model learned. A clothing colour model is also learned, using an image patch directly below the face. These colours are then used when projecting a generative 3D body model, which is compared to the thresholded body. 

\citet{Deutscher00} have proposed the use of simulated annealing to solve the high dimensional search problem associated with 3D pose estimation. Here, a set of weighting functions are used to drive the particle filter search to possible solutions. \citet{Davison2001} perform 3D tracking using multiple cameras and a simulated annealing search. In this case, generative body models are used to create edge and foreground templates, which are compared to those observed using a sum of squared distances metric.

The difficulties in 3D pose estimation from 2D images have led some researchers to focus on 2D pose estimation in images, a slightly better posed problem. \citet{Hua05} apply Markov chain Monte Carlo estimation to fit a set of 2D quadrangles to humans in images, using an observation model combining colour measurements of the head and hands (learned after face detection), and line segments extracted from the torso. 

Applying Monte Carlo search techniques to pose estimation has the benefit of allowing a number of constraints and priors to be incorporated. However, the large number of constraints and complex models required to direct the high dimensional search is hardly ideal, and somewhat inelegant, resulting in large processing burdens. The incorporation of these constraints through priors obtained directly from training data is proposed here, in an attempt to simplify the sampling stages. 

The process of learning constraints from training data has been advocated by \citet{Yu2013}, who clustered 3D body positions according to various action categories, then used action recognition and 2D body parts detected using a deformable part model to predict 3D pose with a random forest. The use of action recognition restricts the possible pose search space, allowing for faster and more accurate pose estimation.

\citet{Howe99} used 3D motion capture data to train a Gaussian mixture model prior, which when combined with a Gaussian error model of 2D tracked body parts allows a 3D pose estimate to be computed using Expectation maximisation. Our approach is similar to this as it also uses a Gaussian mixture prior to incorporate body constraints, but differs through the inclusion of temporal motion tracking using recursive Bayesian estimation. In addition, only a subset of body parts need to be detected for body tracking. A description of our pose estimation method follows.

\section{Bayesian filtering for human pose estimation}
\label{sec:Bayes}

Assuming the human body can be modelled as an unobserved Markov process with a set of joint states $\mathbf{x}_t$ at time $t$, recursive Bayesian estimation allows states to be updated as measurements $\mathbf{z}_t$ are made.
\begin{eqnarray}
	p\left(\mathbf{x}_t|\mathbf{z}_{1:t-1}\right) &=& \int{p\left(\mathbf{x}_t|\mathbf{x}_{t-1}\right) p\left(\mathbf{x}_{t-1}|\mathbf{z}_{1:t-1}\right)\text{d}\mathbf{x}_{t-1}} \label{eq1} \\
	p\left(\mathbf{x}_t|\mathbf{z}_{1:t}\right) &=& \eta p\left(\mathbf{z}_{t}|\mathbf{x}_{t}\right)p\left(\mathbf{x}_{t}|\mathbf{z}_{1:t-1}\right) 
\end{eqnarray}
Here, $\eta$ is a normalising constant and the nomenclature $\mathbf{x}_{1:t}$ refers to the collection of states from time step 1 to t. This process allows for continual state estimation that includes temporal information, using a transition model to predict state changes and an observation model to introduce measurement information.

For human body tracking, the state vector $\mathbf{x}_t$ could comprise the 3D positions of all joints of interest, camera position and orientation, but this causes a number of estimation difficulties when only 2D image measurements obtained from a single camera measurements are available. In this case, image measurements are a non-linear function of the camera position and orientation, which complicates the tracking problem significantly.

This complication can be avoided by performing all filtering in the image plane and only returning to 3D coordinates when a state estimate is obtained. Let $u/\lambda$ and $v/\lambda$ be image coordinates of a body joint, $[X,Y,Z]$, observed by a camera with 6 degree-of-freedom pose $[t_x,t_y,t_z,\alpha,\beta,\gamma]$, 
\begin{eqnarray}
\lambda \begin{bmatrix} u/\lambda\\ v/\lambda\\ 1 \end{bmatrix} &=& \mathbf{K} \begin{bmatrix}
\begin{bmatrix} \cos{\gamma} &-\sin{\gamma} & 0\\ \sin{\gamma} & \cos{\gamma} & 0\\ 0 & 0 & 1\end{bmatrix}
\begin{bmatrix} 1 & 0 & 0\\ 0 &\cos{\beta} &-\sin{\beta}\\ 0 & \sin{\beta} & \cos{\beta}\end{bmatrix}
\begin{bmatrix} \cos{\alpha} &0 & \sin{\alpha}\\ 0 & 1 & 0\\-\sin{\alpha} & 0 & \cos{\alpha}\end{bmatrix} & \begin{bmatrix}t_x\\ t_y\\ t_z \end{bmatrix}\end{bmatrix} \begin{bmatrix} X\\ Y\\ Z\\ 1\end{bmatrix},\nonumber \\
\begin{bmatrix} u\\ v\\ \lambda \end{bmatrix} &=& \begin{bmatrix}\mathbf{\bar{p}}_1 &\mathbf{\bar{p}}_2 &\mathbf{\bar{p}}_3 &\mathbf{\bar{p}}_4\end{bmatrix} \begin{bmatrix} X\\ Y\\ Z\\ 1\end{bmatrix}.\label{projEq}
\end{eqnarray}
Here, $\mathbf{K}$ denotes an intrinsic camera calibration matrix,
\begin{equation}
\mathbf{K} = \begin{bmatrix} f_x & 0 & c_x\\ 0 & f_y & c_y\\ 0 & 0 & 1\end{bmatrix},
\end{equation}
with $f_x$ and $f_y$ focal distances and $c_x, c_y$ coordinates of the camera's principal point. 

Selecting a state vector comprising the scale parameter $\lambda$, image plane coordinates $u/\lambda$, $v/\lambda$ and camera pose allows us to make direct comparisons between state and measurements. Once a state estimate is made, returning to 3D coordinates is trivial, with
\begin{equation}
\begin{bmatrix}X\\Y\\Z\end{bmatrix} = \begin{bmatrix}\mathbf{\bar{p}}_1 &\mathbf{\bar{p}}_2 &\mathbf{\bar{p}}_3\end{bmatrix}^{-1}\left(\begin{bmatrix} u\\ v\\ \lambda \end{bmatrix} - \mathbf{\bar{p}}_4 \right), \label{3Dest}
\end{equation}
and $\mathbf{p}_j$ denoting the $j$-th column vector of the projection matrix in (\ref{projEq}).

\subsection{Transition model}

\label{transsec}

We construct a transition model by combining a simple motion model with an objective function or prior:
\begin{equation}
p\left(\mathbf{x}_t|\mathbf{x}_{t-1}\right) = \frac{\hat{p}\left(\mathbf{x}_{t}|\mathbf{x}_{t-1}\right) \Phi\left(\mathbf{x}_t\right)}{\int{\hat{p}\left(\mathbf{x}_{t}|\mathbf{x}_{t-1}\right)\Phi\left(\mathbf{x}_t\right)\text{d}\mathbf{x}_t}}.
\end{equation}

\begin{figure}[!ht]
	\centering
	\includegraphics[width=0.5\textwidth]{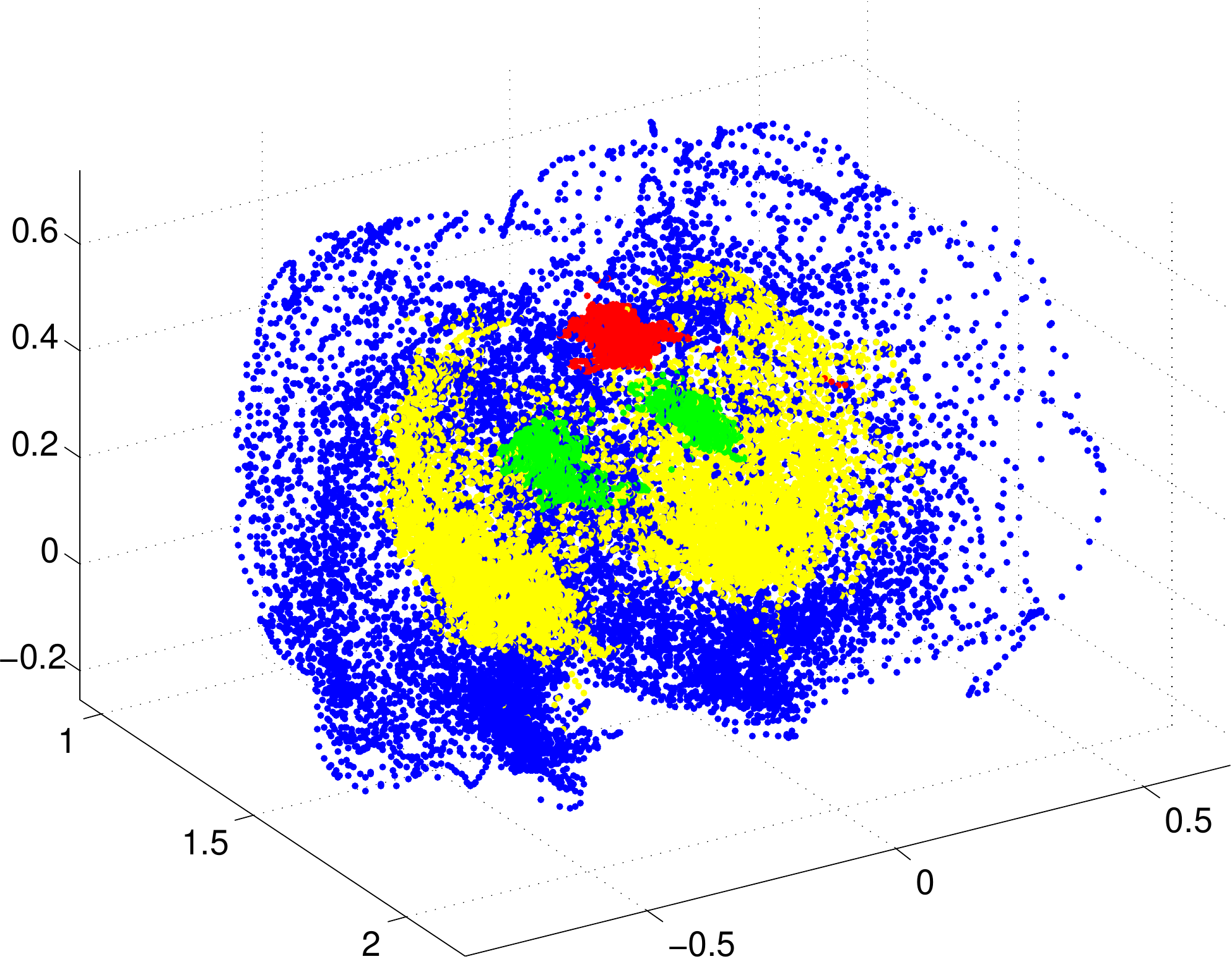}
	\caption[Upper body joint distributions in 3D]{3D upper body joint distributions are captured by recording a human upper body as it undergoes common motions. (Head - red, shoulders - green, elbows - yellow, hands - blue)}
	\label{poseData3D}
\end{figure}

\begin{figure}[!ht]
	\centering
	\subfloat[Head distribution]{\begin{overpic}[width=0.25\textwidth]{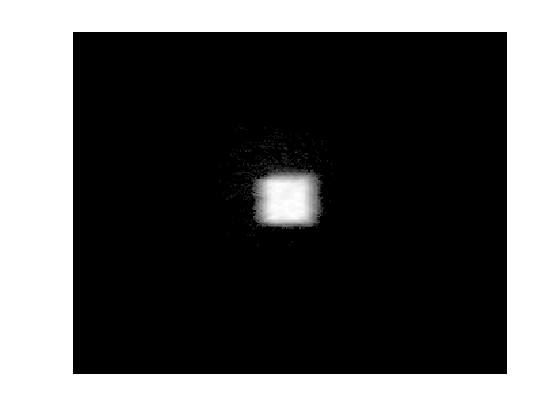}\label{fig:head}\end{overpic}}
	\subfloat[Shoulder distributions]{\begin{overpic}[width=0.25\textwidth]{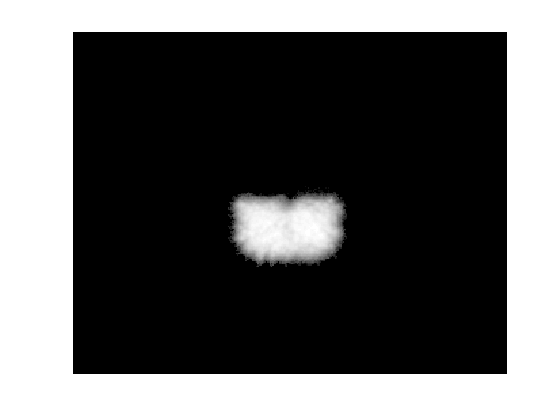}\label{fig:shoulders}\end{overpic}}
	\subfloat[Elbow distributions]{\begin{overpic}[width=0.25\textwidth]{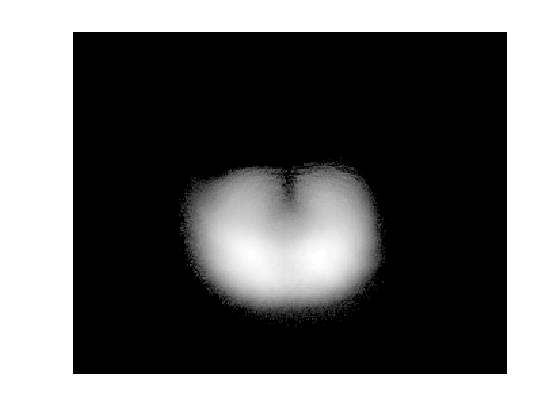}\label{fig:elbows}\end{overpic}}
	\subfloat[Hand distributions]{\begin{overpic}[width=0.25\textwidth]{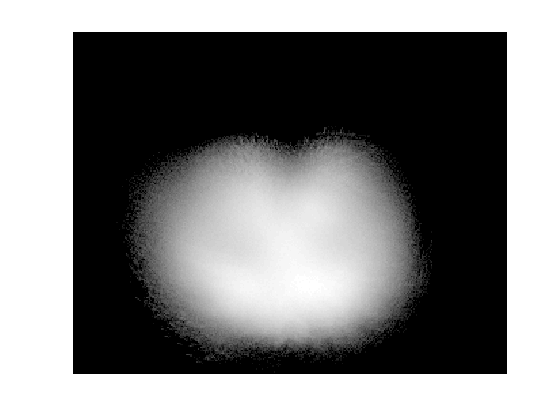}\label{fig:hands}\end{overpic}}
	\caption[Upper body joint distributions in 2D]{Upper body joint distributions are projected into 2D over a range of viewpoints to generate 2D joint position distributions (A limited range of viewpoints are used for illustration to allow for greater clarity). Lighter colours indicate more likely positions.}
	\label{poseData2D}
\end{figure}

This decomposition is useful as it allows a prior distribution covering the probability of a human pose occurring to be used to incorporate likely human poses into the motion model. This distribution is obtained offline, by fitting a Gaussian mixture model (GMM) to a large dataset of recorded human body poses. The positions of upper body joints of interest are tracked using a Kinect sensor \citep{xboxtracker}. Recorded 3D joint positions are then projected into 2D, assuming a pinhole camera with a known camera calibration matrix, $\mathbf{K}$, and a random set of camera viewpoints within a set of constraints ($|\lambda|$, $|\beta|$ and $|\alpha|$ $\le 30^\circ$; $|t_x|,|t_y|$ and $|t_z|$ translation $\le 0.5$ m). This provides a much larger set of recorded 2D joint positions. Figure \ref{poseData3D} shows the original 3D recorded pose data, and the corresponding 2D pose data generated through the synthetic viewpoints is shown in Figure \ref{poseData2D}.

This large dataset is infeasible to work with, and so the Gaussian mixture model of this distribution is a useful form of dimension reduction. A more detailed description on GMMs and their training is provided in Appendix \ref{GMMTraining}. The Gaussian mixture model is denoted by $\Phi\left(\mathbf{x}_t\right)$, the probability of an upper body pose $\mathbf{x}_t$ occurring,
\begin{equation}
\Phi\left(\mathbf{x}_{t}\right) =  \sum_{i=0}^{N-1} \pi_i \mathcal{N}\left(\mathbf{x}_{t}|\boldsymbol\mu_i,\mathbf{\Sigma}_i\right) \label{gmm}.
\end{equation}
Learning the GMM can be computationally intensive and a large number of mixture components may be required. This is remedied by assuming independent left and right arms, and training two mixture models instead.

It is unlikely that states will vary much between time steps, and so we use random walk to describe the motion between states: $\mathbf{x}_{t} = \mathbf{x}_{t-1} + \boldsymbol\epsilon$, with $\boldsymbol\epsilon\sim \mathcal{N}\left(\mathbf{0},\mathbf{Q}\right)$, or
\begin{equation}
\hat{p}\left(\mathbf{x}_{t}|\mathbf{x}_{t-1}\right) = \mathcal{N}\left(\mathbf{x}_{t}|\mathbf{x}_{t-1},\mathbf{Q}\right).\label{trans1}
\end{equation}
The covariance matrix $\mathbf{Q}$ in (\ref{trans1}) is assumed to be a diagonal matrix with each diagonal term selected empirically with image dimensions in mind.

The prior learned from the training data inherently contains kinematic constraints, as well as information on more commonly observed poses. It is also extremely compact and simple. Using this prior, and the fact that the product of two multivariate normal densities over random variable $\mathbf{x}$ is another multivariate normal and scaling constant, we can write
\begin{equation}
\hat{p}\left(\mathbf{x}_{t}|\mathbf{x}_{t-1}\right) \Phi\left(\mathbf{x}_t\right) = \sum_{i=0}^{N-1} \pi_i c_i \mathcal{N}\left(\mathbf{x}_{t}|\boldsymbol\mu^i_c,\mathbf{\Sigma}^i_c\right)
\end{equation}
where
\begin{equation}
c_i = \mathcal{N}\left(\mathbf{x}_{t-1}|\boldsymbol\mu_i, \mathbf{Q} + \mathbf{\Sigma}_i\right)
\end{equation}
and
\begin{eqnarray}
\mathbf{\Sigma}^i_c &=& \left(\mathbf{Q}^{-1} + \mathbf{\Sigma_i}^{-1}\right)^{-1},\\
\boldsymbol\mu^i_c &=& \left(\mathbf{Q}^{-1} + \mathbf{\Sigma_i}^{-1}\right)^{-1}\left(\mathbf{Q}^{-1}\mathbf{x}_{t-1} + \mathbf{\Sigma_i}^{-1}\boldsymbol\mu_i\right).
\end{eqnarray}

As a result, the evidence can be computed as
\begin{eqnarray}
\int{\hat{p}\left(\mathbf{x}_{t}|\mathbf{x}_{t-1}\right)\Phi\left(\mathbf{x}_t\right)\text{d}\mathbf{x}_t} &=& \sum_{i=0}^{N-1} \pi_i c_i \int{ \mathcal{N}\left(\mathbf{x}_{t}|\boldsymbol\mu^i_c,\mathbf{\Sigma}^i_c\right) \text{d}\mathbf{x}_t} \nonumber\\
 &=& \sum_{i=0}^{N-1} \pi_i c_i
\end{eqnarray}

This provides the final transition model
\begin{equation}
p\left(\mathbf{x}_t|\mathbf{x}_{t-1}\right) = \frac{\sum_{i=0}^{N-1} \pi_i c_i \mathcal{N}\left(\mathbf{x}_{t}|\boldsymbol\mu^i_c,\mathbf{\Sigma}^i_c\right)}{\sum_{i=0}^{N-1} \pi_i c_i} \label{transition}.
\end{equation}

This model can be viewed as a mixture of discrete Ornstein-Uhlenbeck processes, in that states behave as random walk, but drift towards a set of typically observed mean poses.

\subsection{Observation model}
\label{obssec}

The observation model used here is assumed to be a Gaussian centred about the difference between a subset of states and measurements,
\begin{equation}
p\left(\mathbf{z}_t|\mathbf{x}_t\right) = \mathcal{N}\left(\mathbf{z}_{t}|\mathbf{H}\mathbf{x}_t,\mathbf{R}\right) \label{observation}.
\end{equation}

The pose state contains the image positions of the head, neck, shoulders, elbows and hands, but it is assumed that only the head, neck and hand states can be measured. These measurements correspond to the subset of states used in the measurement model of (\ref{observation}), selected using $\mathbf{H}$. The covariance matrix $\mathbf{R}$ is assumed to be a diagonal matrix with empirically selected diagonal terms, corresponding to a maximum measurement error in pixels, selected with image dimensions in mind. 

\subsection{Particle filter approximation}

An analytical solution to the integral in (\ref{eq1}) is not always easily computed and often an approximation is required. One way of performing this is to approximate the target distribution using a discrete set of $N_s$ samples. Let
\begin{equation}
p\left(\mathbf{x}_{t}|\mathbf{z}_{1:t}\right) \approx \sum_{k=1}^{N_s} w^{k}_{t}\delta\left(\mathbf{x}_{t} - \mathbf{x}^{k}_{t}\right),
\end{equation}

where the weights $w^{k}_{t}$ are chosen using importance sampling. Consider the full posterior distribution over all states and measurements, with initial estimate $\mathbf{x}_0$,
\begin{eqnarray}
p\left(\mathbf{x}_{0:t}|\mathbf{z}_{1:t}\right) &=& \frac{p\left(\mathbf{z}_{1:t}|\mathbf{x}_{0:t}\right) p\left(\mathbf{x}_{0:t}\right)}{p\left(\mathbf{z}_{1:t}\right)} \nonumber\\
&=& \frac{p\left(\mathbf{z}_{t}|\mathbf{x}_{0:t},\mathbf{z}_{1:t-1}\right) p\left(\mathbf{z}_{1:t-1}|\mathbf{x}_{0:t}\right) p\left(\mathbf{x}_{0:t}\right)}{p\left(\mathbf{z}_{t}|\mathbf{z}_{1:t-1}\right)p\left(\mathbf{z}_{1:t-1}\right)} \nonumber\\
&=&  \frac{p\left(\mathbf{z}_{t}|\mathbf{x}_{0:t},\mathbf{z}_{1:t-1}\right) p\left(\mathbf{x}_{0:t}|\mathbf{z}_{1:t-1}\right)}{p\left(\mathbf{z}_{t}|\mathbf{z}_{1:t-1}\right)} \nonumber\\
&\propto& p\left(\mathbf{z}_{t}|\mathbf{x}_{0:t},\mathbf{z}_{1:t-1}\right) p\left(\mathbf{x}_{0:t}|\mathbf{z}_{1:t-1}\right).
\end{eqnarray}

For a Markov process, the current measurement is only dependent on the current state and the current state is only dependent on the previous state, so we can write
\begin{eqnarray}
p\left(\mathbf{z}_{t}|\mathbf{x}_{0:t},\mathbf{z}_{1:t-1}\right) p\left(\mathbf{x}_{0:t}|\mathbf{z}_{1:t-1}\right) &=& p\left(\mathbf{z}_{t}|\mathbf{x}_{t}\right) p\left(\mathbf{x}_{t}|\mathbf{x}_{0:t-1},\mathbf{z}_{1:t-1}\right) p\left(\mathbf{x}_{0:t-1}|\mathbf{z}_{1:t-1}\right), \nonumber \\
&=& p\left(\mathbf{z}_{t}|\mathbf{x}_{t}\right) p\left(\mathbf{x}_{t}|\mathbf{x}_{t-1}\right) p\left(\mathbf{x}_{0:t-1}|\mathbf{z}_{1:t-1}\right).
\end{eqnarray}
Constructing an importance density $q\left(\mathbf{x}_{0:t}|\mathbf{z}_{1:t}\right)$ from which state samples $\mathbf{x}^k_{0:t}$ are easily sampled provides importance weights
\begin{equation}
w^{k}_{t} \propto  \frac{p\left(\mathbf{z}_t|\mathbf{x}^{k}_t\right)p\left(\mathbf{x}^{k}_t|\mathbf{x}^{k}_{t-1}\right) p\left(\mathbf{x}^k_{0:t-1}|\mathbf{z}_{1:t-1}\right)}{q\left(\mathbf{x}^{k}_{0:t}|\mathbf{z}_{1:t}\right)},
\end{equation}
which can be written recursively as
\begin{equation}
w^{k}_{t} \propto  w^{k}_{t-1}\frac{p\left(\mathbf{z}_t|\mathbf{x}^{k}_t\right)p\left(\mathbf{x}^{k}_t|\mathbf{x}^{k}_{t-1}\right)}{q\left(\mathbf{x}^{k}_{t}|\mathbf{x}^{k}_{0:t-1},\mathbf{z}_{1:t}\right)}.
\end{equation}
Since we are only interested in the state at time $t$, and desire an approximation to the density $p\left(\mathbf{x}_{t}|\mathbf{z}_{1:t}\right)$, we can discard the state history and the weight update equation becomes
\begin{equation}
w^{k}_{t} \propto  w^{k}_{t-1}\frac{p\left(\mathbf{z}_t|\mathbf{x}^{k}_t\right)p\left(\mathbf{x}^{k}_t|\mathbf{x}^{k}_{t-1}\right)}{q\left(\mathbf{x}^{k}_{t}|\mathbf{x}^{k}_{t-1},\mathbf{z}_{t}\right)}.
\end{equation}
Unfortunately, sequential importance sampling often suffers from degeneracy problems \citep{Doucet00}, where the weights of most particles become negligible after a few iterations. This is remedied by resampling, which generates a new set of particles by sampling with replacement according to the importance weights. This typically eliminates particles that have small weights and adds emphasis to those with larger importance. Special care needs to be taken as to the selection of the proposal density $q\left(\mathbf{x}^{k}_{t}|\mathbf{x}^{k}_{t-1},\mathbf{z}_{t}\right)$. Ideally this should be as close to the target density as possible. 

The sampling importance resampling (SIR) or bootstrap filter, discussed in detail by \citet{Ristic04}, is frequently used for recursive Bayesian filtering. Here, the importance density is usually chosen to be equal to the transition density,
\begin{equation}
q\left(\mathbf{x}_{t}|\mathbf{x}_{t-1},\mathbf{z}_{t}\right) =  p\left(\mathbf{x}_t|\mathbf{x}_{t-1}\right).
\end{equation}
This reduces the importance weight calculation to 
\begin{equation}
w^{k}_{t} \propto  w^{k}_{t-1} p\left(\mathbf{z}_t|\mathbf{x}^{k}_t\right) \label{weightupdate1}.
\end{equation}
By applying resampling at each time step, the weights become uniform, and the weight update simplifies to
\begin{equation}
w^{k}_{t} \propto  p\left(\mathbf{z}_t|\mathbf{x}^{k}_t\right).
\end{equation}
The SIR filtering procedure is described in more detail in Algorithm \ref{pfalg}.
\begin{algorithm}
	\caption{Sampling importance resampling particle filter}
	\begin{algorithmic}
	\LOOP
		\FOR {$k = 1$ to $N_s$}
			\STATE Draw $\mathbf{x}_t^k \sim p\left(\mathbf{x}_t|\mathbf{x}_{t-1}\right)$ $\rightarrow$ Equation (\ref{transition})
			\STATE $w^{k}$ = $p\left(\mathbf{z}_t|\mathbf{x}_t^{k}\right)$ $\rightarrow$ Equation (\ref{observation})
		\ENDFOR
		\STATE Normalise weights $\mathbf{w} = [w^{1} \hdots w^{N_s}]$ 
		\STATE Resample $\mathbf{x}_t$ according to $\mathbf{w}$
	\ENDLOOP
	\end{algorithmic}
	\label{pfalg}
\end{algorithm}

Resampling may be computationally expensive, so in practise it is not desirable to resample on each iteration. Instead, resampling need only occur when the effective number of particles is below a certain threshold, and the particle filter is close to degeneracy. An estimate of the effective number of particles used by a particle filter \citep{Kong94} is
\begin{equation}
\hat{N}_{eff} = \frac{1}{\sum_{k=1}^{N_s} \left(w_t^k\right)^2}.
\end{equation}

Unfortunately, drawing samples from the Gaussian mixture model of (\ref{transition}) is rather computationally intensive. Sampling from this GMM requires $N_s$ draws from a uniform distribution to select a mixture component according to the model's mixture weights, and a further $N_s$ draws from different Gaussians (due to the dependence of (\ref{transition}) on previous states) to select particles. As an alternative solution, we propose that samples be drawn from the far simpler density $\hat{p}\left(\mathbf{x}_{t}|\mathbf{x}_{t-1}\right)$, which results in the weight update equation of 
\begin{equation}
w^{k}_{t} \propto  \frac{p\left(\mathbf{z}_t|\mathbf{x}^{k}_t\right) \Phi\left(\mathbf{x}^{k}_t\right)}{\left[\int{\hat{p}\left(\mathbf{x}_{t}|\mathbf{x}_{t-1}\right)\Phi\left(\mathbf{x}_t\right)\text{d}\mathbf{x}_t}\right]_{\mathbf{x}^k_{t-1}}}. \label{simplesampler}
\end{equation}

An even more efficient approximation could neglect the scaling term $\left[\int{\hat{p}\left(\mathbf{x}_{t}|\mathbf{x}_{t-1}\right)\Phi\left(\mathbf{x}_t\right)\text{d}\mathbf{x}_t}\right]_{\mathbf{x}^k_{t-1}}$ entirely, although this could potentially introduce evidence bias in the tails of the distribution. In the following section, we will show that ignoring this term is effectively equivalent to modifying the transition model such that a random walk is applied to each mixture component independently, as opposed to the entire distribution.

Particle filter tracking in high dimensions typically relies on good initial particle estimates. In an attempt to remedy this, we start with much larger joint variance along the diagonals of $\mathbf{Q}$ in (\ref{trans1}) and slowly reduce this over a burn-in period, to allow for an initial particle convergence phase. This can be considered a form of simulated annealing, which has been used previously for pose tracking by \citet{Deutscher00}.

\subsection{Mixture Kalman filter}

The particle filter is a useful approximation when dealing with complex probability distributions, which cannot be analytically integrated. However, the use of a Gaussian mixture model in the transition density and a conjugate Gaussian observation model allows us to Rao-Blackwellise the particle filter by performing integrations optimally using a number of Kalman filters to track mixture components, in a manner similar to that described by \citet{alspach1972}. This approach, termed the mixture Kalman filter, has been applied to a number of conditionally linear dynamic models by \citep{chen2000} and \citep{Doucet00}. 

Our goal is to calculate the posterior distribution, $p\left(\mathbf{x}_t|\mathbf{z}_{1:t}\right)$, given a sequence of measurements. Recall that a prior model on human pose, learned from Kinect training data, can be denoted by a weighted summation of Gaussians, with means and variances $\boldsymbol{\mu}_i$ and $\mathbf{\Sigma}_i$ respectively,
\begin{equation}
\Phi\left(\mathbf{x}_t\right) = \sum_{i=1}^N \pi_i \mathcal{N}\left(\mathbf{x}_t|\boldsymbol{\mu}_i,\mathbf{\Sigma}_i\right).
\end{equation}
This distribution can be partitioned if we introduce an indicator variable $i$, which refers to the $i$-th mixture component in the distribution. Then the prior probability over states can be denoted as
\begin{equation}
\Phi\left(\mathbf{x}_t\right) = \sum_{i=1}^N p\left(i\right) \Phi\left(\mathbf{x}_t|i\right)
\end{equation}
with
\begin{eqnarray}
p\left(i\right) &=& \pi_i, \\  \Phi\left(\mathbf{x}_t|i\right) &=& \mathcal{N}\left(\mathbf{x}_t|\boldsymbol{\mu}_i,\mathbf{\Sigma}_i\right).
\end{eqnarray}

Applying the random walk transition density selected in (\ref{trans1}) to each mixture component independently provides the transition density for the body pose conditioned on the indicator variable and previous state,
\begin{eqnarray}
p\left(\mathbf{x}_t|\mathbf{x}_{t-1},i\right) &=& \frac{\hat{p}\left(\mathbf{x}_{t}|\mathbf{x}_{t-1},i\right)\Phi\left(\mathbf{x}_{t}|i\right)}{\int{\hat{p}\left(\mathbf{x}_{t}|\mathbf{x}_{t-1},i\right)\Phi\left(\mathbf{x}_{t}|i\right) \text{d}\mathbf{x}_t}} \nonumber \\
&=& \frac{\mathcal{N}\left(\mathbf{x}_{t}|\mathbf{x}_{t-1},\mathbf{Q}\right) \mathcal{N}\left(\mathbf{x}_t|\boldsymbol{\mu}_i,\mathbf{\Sigma}_i\right)}{\int{\mathcal{N}\left(\mathbf{x}_{t}|\mathbf{x}_{t-1},\mathbf{Q}\right)\mathcal{N}\left(\mathbf{x}_t|\boldsymbol{\mu}_i,\mathbf{\Sigma}_i\right)\text{d}\mathbf{x}_t}},
\end{eqnarray}
which can be solved analytically to provide the normal distribution
\begin{equation}
p\left(\mathbf{x}_t|\mathbf{x}_{t-1},i\right) = \mathcal{N}\left(\mathbf{x}_{t}\bigg|\left(\mathbf{\Sigma}_i^{-1}+\mathbf{Q}^{-1}\right)^{-1}\left(\mathbf{\Sigma}_i^{-1}\boldsymbol{\mu}_i + \mathbf{Q}^{-1}\mathbf{x}_{t-1}\right), \left(\mathbf{\Sigma}_i^{-1}+\mathbf{Q}^{-1}\right)^{-1}\right). \label{motionKF}
\end{equation}

Assuming only a subset of states, $\mathbf{z}_t = \mathbf{H}\mathbf{x}_t$, can be observed in the presence of zero-mean Gaussian measurement noise with covariance $\mathbf{R}$ provides a measurement model,
\begin{equation}
p\left(\mathbf{z}_t|\mathbf{x}_t\right) = \mathcal{N}\left(\mathbf{z}_{t}\bigg|\mathbf{H}\mathbf{x}_t,\mathbf{R}\right). \label{measureKF}
\end{equation}

Equations (\ref{motionKF}) and (\ref{measureKF}) are of the form required for optimal Bayesian filtering using the Kalman filter \citep{Kalman60}. The Kalman filter marginalises out historical states and provides the posterior distribution of a state for a given trajectory of indicator variables,  $p\left(\mathbf{x}_t|\mathbf{z}_{1:t},\boldsymbol\lambda^j_t\right)$, conditioned on a mixture component. Here, the boldface $\boldsymbol\lambda^j_t=\left[\lambda_1,\lambda_2,\hdots,\lambda_t = i\right]$, with $i \in [1,N]$ is used to denote the $j$-th trajectory of mixture components, from time steps $1$ to $t$. First, $\mathbf{\hat{x}}_t(\boldsymbol\lambda^j_t)$, a prediction of the state mean conditioned on a particular sequence of indicator variables up to time $t$ is made using the transition model of (\ref{motionKF}), 
\begin{equation}
\mathbf{\hat{x}}_t(\boldsymbol\lambda^j_t) = \mathbf{F}_{\lambda_t}\mathbf{\tilde{x}}_{t-1}(\boldsymbol\lambda^j_{t-1}) + \mathbf{B}_{\lambda_t}\boldsymbol{\mu}_{\lambda_t},
\end{equation}
assuming no process noise, with
\begin{equation}
\mathbf{F}_{\lambda_t} = \left(\mathbf{Q}^{-1} + \mathbf{\Sigma}_{\lambda_t}^{-1}\right)^{-1}\mathbf{Q}^{-1}
\end{equation}
and
\begin{equation}
\mathbf{B}_{\lambda_t} = \left(\mathbf{Q}^{-1} + \mathbf{\Sigma}_{\lambda_t}^{-1}\right)^{-1}\mathbf{\Sigma}_{\lambda_t}^{-1}.
\end{equation}
The existing uncertainty in the mixture component is propagated through the linear process model, and uncertainty in the model included, to provide the predicted mixture covariance, 
\begin{equation}
\mathbf{\hat{P}}_{t}(\boldsymbol\lambda^j_t) = \mathbf{F}_{\lambda_t}\mathbf{\tilde{P}}_{t-1}(\boldsymbol\lambda^j_{t-1})\mathbf{F}_{\lambda_t}^{\text{T}} + \left(\mathbf{Q}^{-1}+\mathbf{\Sigma}_{\lambda_t}^{-1}\right)^{-1}.
\end{equation}
When observations are made, the measurement and covariance residuals are calculated using
\begin{equation}
\mathbf{y}_{t} = \mathbf{z}_t - \mathbf{H}\mathbf{\hat{x}}_t(\boldsymbol\lambda^j_t)
\end{equation}
and 
\begin{equation}
\mathbf{S}_{t} = \mathbf{H}\mathbf{\hat{P}}_{t}(\boldsymbol\lambda^j_t)\mathbf{H}^{T} + \mathbf{R}. 
\end{equation} 
These residuals are then used to provide the updated mean and covariance estimates
\begin{eqnarray}
\mathbf{\tilde{x}}_t(\boldsymbol\lambda^j_t) &=& \mathbf{\hat{x}}_t(\boldsymbol\lambda^j_t) + \mathbf{K}^i_t\mathbf{y}_{t},\\
\mathbf{\tilde{P}}_{t}(\boldsymbol\lambda^j_t) &=& \left(\mathbf{I} - \mathbf{K}_t\mathbf{H}\right)\mathbf{\hat{P}}_{t}(\boldsymbol\lambda^j_t),
\end{eqnarray} 
where $\mathbf{K}_t = \mathbf{\hat{P}}_{t}(\boldsymbol\lambda^j_t)\mathbf{H}^{\text{T}}\mathbf{S}_{t}^{-1}$ is the optimal Kalman gain for a linear system. Finally, the posterior density for the state conditioned on a trajectory of mixture components can then be described by a Gaussian,
\begin{equation}
p\left(\mathbf{x}_t|\mathbf{z}_{1:t},\boldsymbol\lambda^j_t\right) = \mathcal{N}\left(\mathbf{x}_t|\mathbf{\tilde{x}}_{t}(\boldsymbol\lambda^j_t),\mathbf{\tilde{P}}_{t}(\boldsymbol\lambda^j_t)\right)
\end{equation}

Using this information, the probability of an indicator variable trajectory conditioned on the sequence of measurements, $p\left(\boldsymbol\lambda^j_t|\mathbf{z}_{1:t}\right)$, can be used  to obtain the target distribution 
\begin{equation}
p\left(\mathbf{x}_t|\mathbf{z}_{1:t}\right) = \sum^N_{i=1} \sum^M_{j=1}  p\left(\mathbf{x}_t|\mathbf{z}_{1:t},\boldsymbol\lambda^j_t\right)p\left(\boldsymbol\lambda^j_t|\mathbf{z}_{1:t}\right).  \label{indicatorInt}
\end{equation}
Here, $N$ denotes the number of indicator components in the motion model, and $M$ the number of indicator variable trajectories.

The conditional indicator probability is obtained by marginalising the joint state indicator distribution,
\begin{eqnarray}
p\left(\boldsymbol\lambda^j_t|\mathbf{z}_{1:t}\right) &=& \int p\left(\mathbf{x}_t, \boldsymbol\lambda^j_t|\mathbf{z}_{1:t}\right) \text{d}\mathbf{x}_t \nonumber \\
&=& \int {\frac{p\left(\mathbf{z}_{1:t}|\mathbf{x}_t, \boldsymbol\lambda^j_t\right) p\left(\mathbf{x}_t, \boldsymbol\lambda^j_t\right)}{p\left(\mathbf{z}_{1:t}\right)}\text{d}\mathbf{x}_t} \nonumber \\
&=& \int {\frac{p\left(\mathbf{z}_{t}|\mathbf{x}_t\right) p\left(\mathbf{z}_{1:t-1}|\mathbf{x}_t, \boldsymbol\lambda^j_t\right)p\left(\mathbf{x}_t, \boldsymbol\lambda^j_t\right)}{p\left(\mathbf{z}_{1:t}\right)}\text{d}\mathbf{x}_t} \nonumber \\
&=& \int {\frac{p\left(\mathbf{z}_{t}|\mathbf{x}_t\right) p\left(\mathbf{x}_t, \boldsymbol\lambda^j_t|\mathbf{z}_{1:t-1}\right)}{p\left(\mathbf{z}_{t}|\mathbf{z}_{1:t-1}\right)}\text{d}\mathbf{x}_t} \nonumber \\
&=& \int {\frac{p\left(\mathbf{z}_{t}|\mathbf{x}_t\right) p\left(\mathbf{x}_t|\boldsymbol\lambda^j_t, \mathbf{z}_{1:t-1}\right)p\left(\boldsymbol\lambda^j_t|\mathbf{z}_{1:t-1}\right)}{p\left(\mathbf{z}_{t}|\mathbf{z}_{1:t-1}\right)}\text{d}\mathbf{x}_t} \nonumber \\
&\propto& p\left(\lambda^j_t\right) p\left(\boldsymbol\lambda^j_{t-1}|\mathbf{z}_{1:t-1}\right)\int {p\left(\mathbf{z}_{t}|\mathbf{x}_t\right) p\left(\mathbf{x}_t|\boldsymbol\lambda^j_t, \mathbf{z}_{1:t-1}\right)\text{d}\mathbf{x}_t}. \label{factoredIndicator}
\end{eqnarray}
The contents of the integral in (\ref{factoredIndicator}) are known, with $p\left(\mathbf{z}_{t}|\mathbf{x}_t\right)$ the normal measurement model of (\ref{measureKF}) and $p\left(\mathbf{x}_t|\boldsymbol\lambda^j_t, \mathbf{z}_{1:t-1}\right)$ the result of the Kalman filter prediction step, also Gaussian, which we shall denote as $\mathcal{N}\left(\mathbf{x}_{t}|\mathbf{\hat{x}}_t(\boldsymbol\lambda^j_t),\mathbf{\hat{P}}_t(\boldsymbol\lambda^j_t)\right)$. As a result, (\ref{factoredIndicator}) reduces to an iterative form
\begin{equation}
p\left(\boldsymbol\lambda^j_t|\mathbf{z}_{1:t}\right) = \eta \mathcal{N}\left(\mathbf{z}_t\bigg|\mathbf{H}\mathbf{\hat{x}}_t(\boldsymbol\lambda^j_t), \mathbf{H}\mathbf{\hat{P}}_t(\boldsymbol\lambda^j_t)\mathbf{H}^\text{T} + \mathbf{R}\right) p\left(\lambda^j_t\right) p\left(\boldsymbol\lambda^j_{t-1}|\mathbf{z}_{1:t-1}\right),
\end{equation}
with $\eta$ a normalising constant.

Unfortunately the sums in (\ref{indicatorInt}) are hard to compute, as the number of trajectories grows exponentially with each filtering iteration, so in practice we approximate (\ref{indicatorInt}) as a weighted sum of trajectories of interest,
\begin{equation}
p\left(\mathbf{x}_t|\mathbf{z}_{1:t}\right) \approx  \sum_{j=1}^M w_t^{j} p\left(\mathbf{x}_t|\mathbf{z}_{1:t},\boldsymbol\lambda^j_t\right) \label{weightedEstimate}.
\end{equation}

The mixture Kalman filter uses importance sampling to select the subset of trajectories, with weights updated using 
\begin{equation}
w_t^{j} \propto \frac{p\left(\boldsymbol\lambda^j_t|\mathbf{z}_{1:t}\right)}{q\left(\lambda^j_t|\mathbf{z}_{1:t}\right)} = p\left(\mathbf{z}_t\bigg|\mathbf{H}\mathbf{\hat{x}}_t(\boldsymbol\lambda^j_t), \mathbf{H}\mathbf{\hat{P}}_t(\boldsymbol\lambda^j_t)\mathbf{H}^\text{T} + \mathbf{R}\right) w_{t-1}^{j},
\end{equation}
when indicator variables $\lambda^j_t$ are sampled from the proposal density, $q\left(\lambda_t|\mathbf{z}_{1:t}\right) =  p\left(\lambda_t\right)$.

Using the sampled indicator variables and these weights, a maximum a posteriori estimate for the upper body pose can be obtained through a weighted combination of updated mixture means,
\begin{equation}
\bar{\mathbf{x}}_t \approx \sum_{j=1}^M \mathbf{\tilde{x}}_t(\boldsymbol\lambda^j_t) w_t^{j}.
\end{equation}
This pose estimate is easily calculated, typically requiring only a small number of parallel Kalman filters, so is far more efficient than a bootstrap particle filter approximation. Finally, a 3D human body pose is obtained by evaluating (\ref{3Dest}) at the estimated state. 

In practice, many of the weights, $w_t^{j}$, can become negligible after a few iterations, with only a few Gaussians contributing to the final pose estimate. This is remedied by resampling with replacement whenever the effective number of particles falls too low.

Importance sampling can be expensive, so a suboptimal approximation to (\ref{indicatorInt}) could be obtained by selecting a fixed set of trajectories by some other means. A number of mixture reduction schemes \cite{Salmond90},\cite{Blom84} have been proposed previously, but many of these can be expensive. For example, trajectories could be selected by performing the update step for each possible mixture component and input trajectory, then discarding trajectories with low indicator weights. This approach is termed the split-track filter \cite{Smith78}. We propose that a subset of trajectories be selected by only retaining trajectories where $\lambda^j_t = \lambda^j_{t-1}$ and $\lambda^j_1 = j$, which forces continuity between indicator variables and guarantees that every mixture component is fairly represented in the posterior distribution, in effect giving more weight to the prior distribution on human poses. Here, weights are updated using
\begin{equation}
w_t^{j} \propto  p\left(\lambda^j_t\right) p\left(\mathbf{z}_t\bigg|\mathbf{H}\mathbf{\hat{x}}_t(\boldsymbol\lambda^j_t), \mathbf{H}\mathbf{\hat{P}}_t(\boldsymbol\lambda^j_t)\mathbf{H}^\text{T} + \mathbf{R}\right) w_{t-1}^{j}.
\end{equation}

As mentioned previously, weights can tend to zero for a given mixture component. Resampling in this case is not ideal, as it could become impossible for this mixture to contribute towards the pose estimate regardless of future measurements. This is undesirable as it effectively removes the mean-reverting properties of the process model. This is remedied by adding a small uniform prior, $\epsilon > 0$, to the weights on each iteration. The size of $\epsilon$ controls the speed at which the process model is able to transition between reverting to the different mixture means in the pose prior.


\section{Tracking results}
\label{sec:TrackResults}
In the previous section, we introduced a motion model suitable for upper body tracking using recursive Bayesian estimation and discussed a selection of tracking schemes to perform this. The first, a bootstrap particle filter, makes proposals from the GMM transition model in (\ref{transition}) and uses the weight update equation in (\ref{weightupdate1}). This sampling step is quite time consuming, and the second, faster scheme discussed draws samples from the simple random walk in (\ref{trans1}) for use with the weight update equation in (\ref{simplesampler}). The third tracker neglects the scaling evidence term in the weight update equation of (\ref{simplesampler}) to obtain an even faster approximation. Neglecting this term is equivalent to assuming independence across mixture components, or that the transition noise is added to each mixture component separately. The final two tracking schemes introduced also use this slightly modified transition model, where noise is added to each mixture component independently, to allow for an iterative solution using the mixture Kalman filter. The importance sampling step used to select indicator variables in this scheme can be time consuming, so an approximation using a deterministic set of indicator trajectories was also proposed, where each indicator variable selected is paired with a specific trajectory.

\begin{figure}
	\centering
	\includegraphics{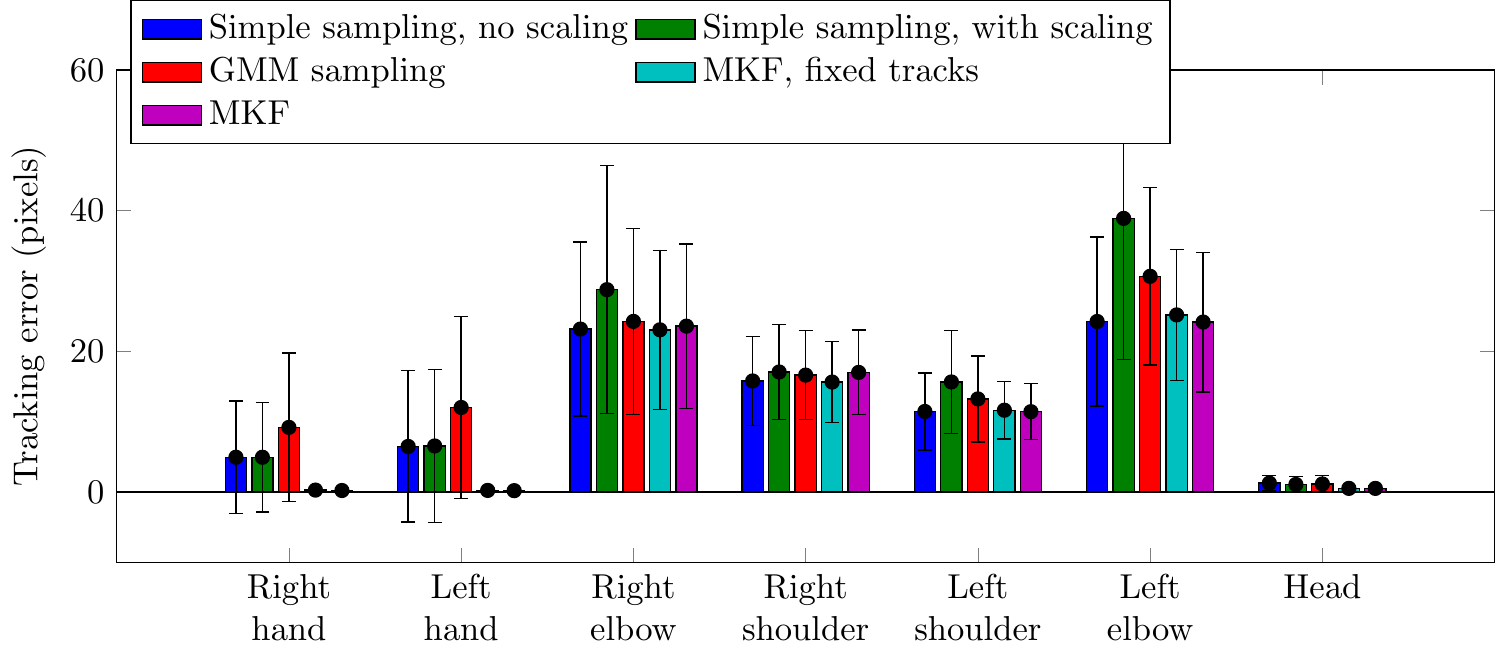}
	\caption[Average joint error]{Average joint error over an image sequence containing a moving person.} \label{avetrackerror}
\end{figure}

Results obtained after applying the five tracking schemes discussed to manually annotated image sequences are provided here. Each of the schemes was applied to image sequences with a moving person, and the pose estimates compared to those obtained using the Kinect motion tracker. Independent datasets were used to learn the pose priors and test the pose estimates. Figure \ref{avetrackerror} shows the mean pixel error for each joint over the test sequence. 

No simulated annealing was used for the scheme sampling from the full Gaussian mixture model, as this required a larger level of noise in the transition model in order to avoid losing track of the joints completely. The figure shows that the best performance was obtained using the mixture Kalman filter (MKF) approaches. Of the particle filter approaches, the sampling scheme with no scaling converged and tracked the actual pose best, with rather poor tracking achieved when weighting was included. The theoretically preferred Gaussian mixture model sampling was unable to adequately track motion, presumably due to its slow convergence. 

\begin{figure}
	\centering
	\includegraphics{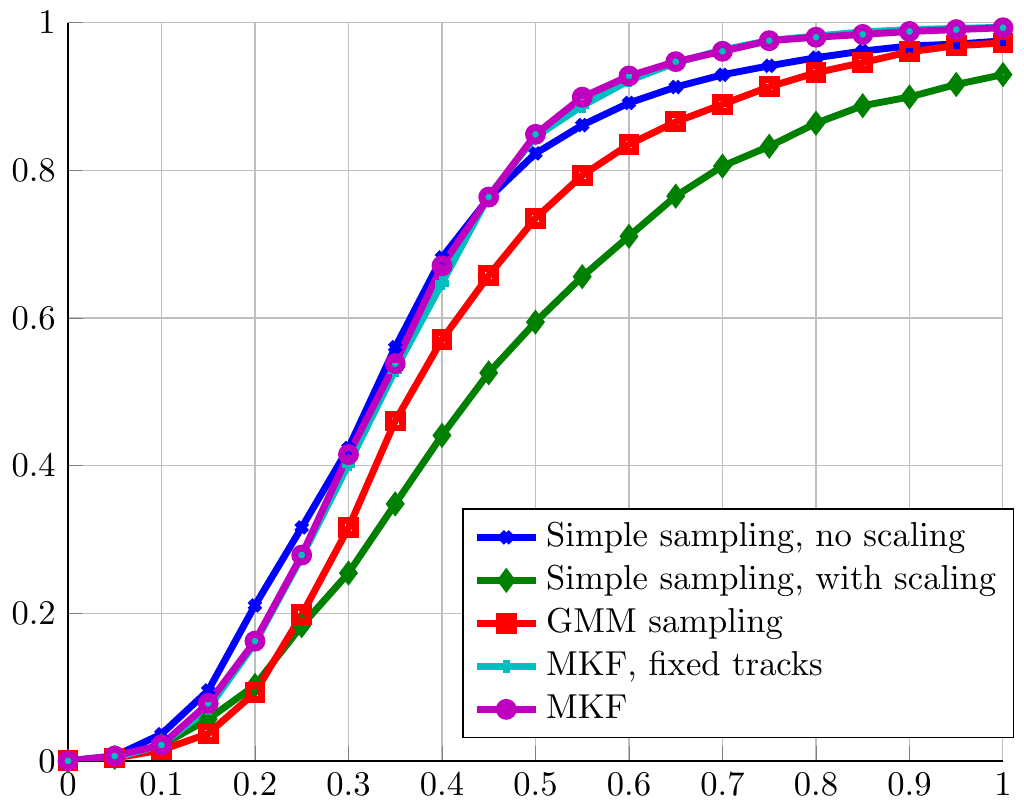}
	\caption{Probability of correct pose (PCP) curves for the various tracking approaches show that the MKF and simple sampling (without scaling) strategies are generally the best estimators of pose.}	\label{fig:pcp}
\end{figure}

A commonly used metric that assesses the performance of 2D pose estimation algorithms is the probability of correct pose (PCP) \citep{Yang2011}, which shows the percentage of correctly localised body parts, where a body part is deemed to be correctly localised if its end points fall within some fraction of the ground truth body part length. Figure \ref{fig:pcp} shows the PCP curves for each of the various tracking schemes (only forearm and upper arm localisation is considered). This metric highlights the performance of the Mixture Kalman filters.

\begin{figure}
	\centering
	\includegraphics{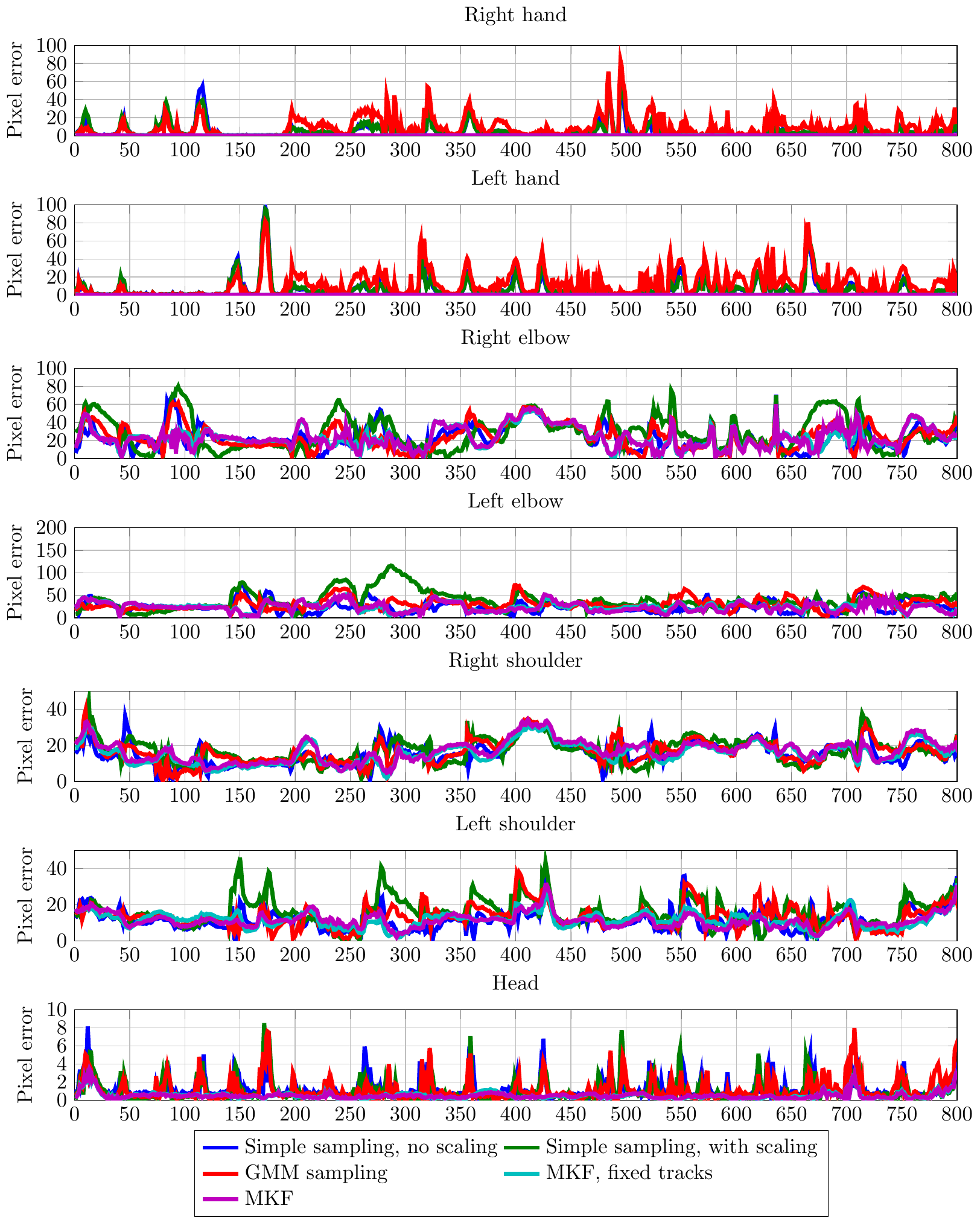}
	\caption[Tracking error for a moving upper body pose]{Tracking error for an image sequence of a moving upper body, when each of the five suggested tracking schemes is applied.} \label{trackerror}
\end{figure}

Figure \ref{trackerror} indicates the pixel errors obtained for each joint over the entire test period. Noticeable error spikes that occur when the particle filters are used are not present in the mixture Kalman filter results. The superior performance of the mixture Kalman filter approaches and the simple sampling scheme disregarding scaling make it is clear that the modified transition density of (\ref{motionKF}), where noise is added to each component independently, is a better model of human motion than that of (\ref{transition}).

Table \ref{tab:iterationTimes} shows the average time taken for each filter iteration, when each of the suggested tracking schemes is used. It is clear that sampling from the full GMM is significantly more time consuming than the simple sampling, but that the mixture Kalman filters are far faster than all of the particle filter approximations.
\begin{table}
	\caption{Average iteration times}
	\label{tab:iterationTimes}
	\centering
	\begin{tabular}{|l|l|}
		\hline
		\bf{Sampling strategy} & \bf{Time} \\
		\hline		\hline
		Simple sampling, with scaling (10000 particles) & 0.046 s\\
		\hline
		Simple sampling, no scaling (10000 particles) & 0.028 s\\
		\hline
		GMM sampling (10000 particles) & 2.947 s\\
		\hline
		MKF (30 mixture components) & 0.021 s\\
		\hline
		MKF, fixed tracks (30 mixture trajectories) & 0.015 s\\
		\hline
	\end{tabular}
\end{table}

\begin{figure}
	\centering
	\includegraphics{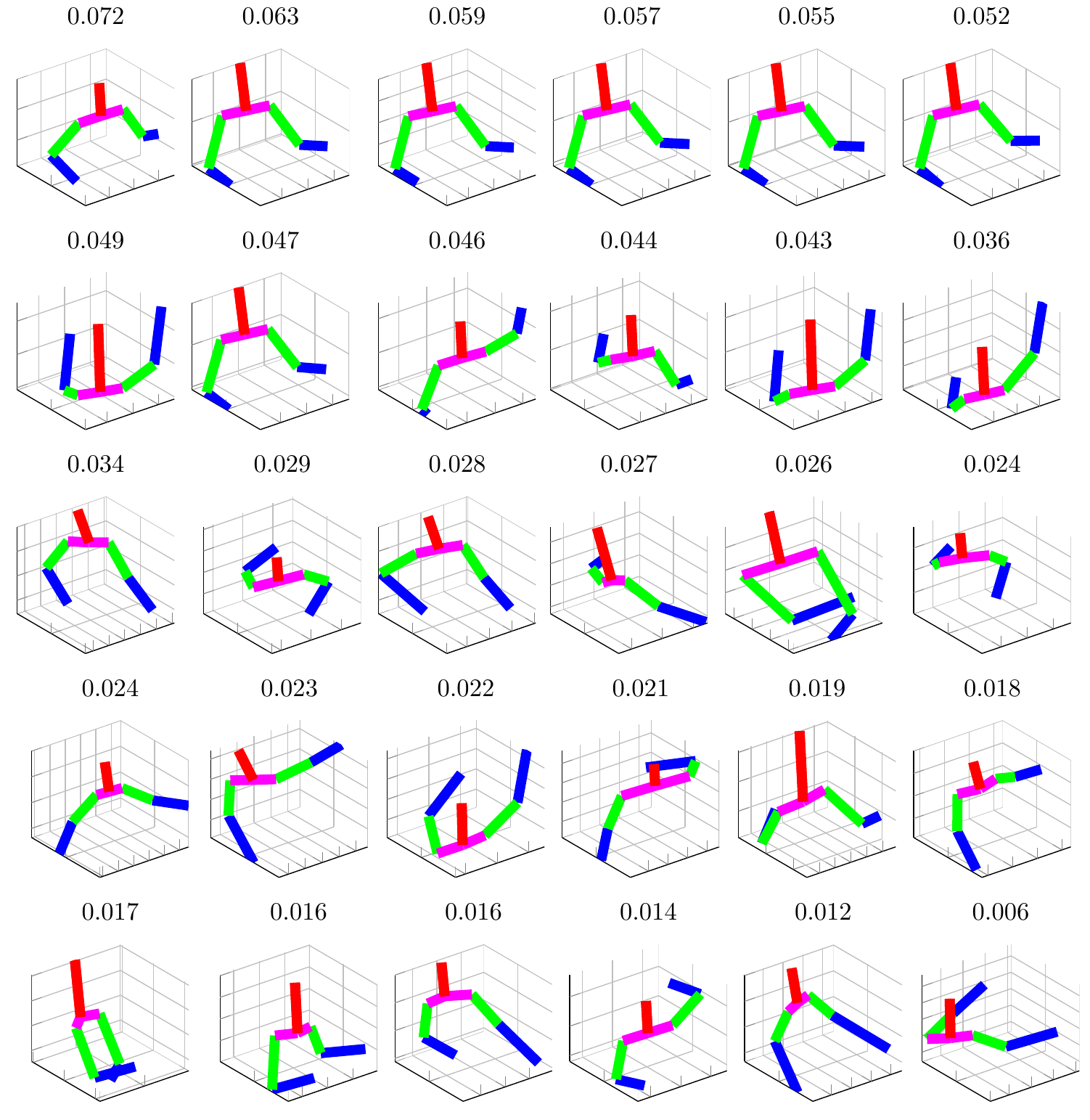}
	\caption{Mean poses and weights learned from the training data show that rare events such as raised and bent arms are considered less important than frequently observed poses by our model. As a result, these poses are less likely to contribute to a pose estimate when importance sampling is used because the corresponding indicator variables are sampled less frequently. Forcing trajectories to contain these indicators by using the proposed deterministic indicator selection means point estimates are more likely to include rare events when they do occur.}\label{fig:poseMeans}
\end{figure}

Note that the mixture Kalman filter approximation using deterministically selected tracks provides almost identical performance to the MKF using sampled indicator variables, but is significantly faster. Qualitative results show that using the MKF with fixed tracks provides a much smoother tracking result (see accompanying videos). This deterministic MKF also appears to be better at dealing with uncommon scenarios such as raised arms (Figure \ref{fig:poseMeans}), which is presumably due to the fact that all mixture components are paired with a specific trajectory, and as a result can always contribute to a pose estimate. In contrast, the original MKF will place emphasis on mixture components that carry more weight, and this effect will propagate until components of less weight become negligible. The MKF distribution obtained when sampling indicators may be closer to the true joint distribution, but appears less suited to providing a point estimate as a result, since it appears to be more susceptible to ambiguities in the pose estimation.


\section{Automatically obtaining measurements}
\label{headhandmeasure}
Thus far, manually annotated images have been used to compare pose estimation schemes. The process of detecting head and hand positions and incorporating these into the filtering framework is now described.

\subsection{Face detection and tracking}
Automatic face detection is frequently required by computer vision systems and a large number of extremely effective algorithms are available to accomplish this. In this work, an OpenCV \citep{OpenCV} implementation of the well known \citet{Viola01} face detector is applied. This detector classifies faces using a cascade of boosted classifiers, trained using the responses to Haar-like features.

The face detector is trained over a wide selection of faces, but only frontal faces are used as positive training examples, in line with our end application of human-robot interaction. In these applications, a robot should only attempt to engage with a person who is looking directly at it, in the same way humans make eye contact when conversing.   

The face detection is augmented through the addition of face tracking using a Kalman filter \citep{Kalman60} and constant velocity motion model, applying a modified version of the simple object tracker described by \citet{Burke10}. This tracking provides a degree of robustness to false negatives (faces present, but not detected), and can be used to reject false positives (faces detected, but not present) as these tend to be detected sporadically and fail to provide lasting tracks.





%

With each input image, detected faces are compared to tracked faces using a Euclidean norm distance measure, including the position and size (height and width) of the faces. If this measure falls below a certain threshold, the update stage of the Kalman filter is applied to the corresponding tracked face. If this is not the case, a new track is started. When faces have not been observed for a certain number of time steps, they are removed from the list of tracked faces. Similarly, tracked faces are only used if the track has lasted for a predefined length of time.

\subsection{Hand detection and tracking}

Once detected, faces contain important information, which can assist in the detection of other body parts. This section shows how the detected face can be used to determine the tracked person's skin colour, and segment hands. 

First, a histogram of the colours (Lab colour space) present in a square image patch bounding the detected face is back-projected to provide a likelihood map of image areas resembling skin. Here, back-projection refers to the process of evaluating the probability of an image pixel being skin coloured, with the likelihood approximated by a histogram of the detected pixel values in a training image patch. An exponentially weighted moving average filter favouring historical measurements is applied to the histogram to limit the effects of spurious lighting dependent observations.

Originally, a Gaussian mixture model was trained using this image patch and used for skin colour segmentation, but this proved computationally expensive, and provided little improvement over a simple back-projection. In order to assist in the recognition of hands, areas of high likelihood are only labelled as left or right hands when placed within an initialisation area, consisting of the left and right halves of the input image. This serves as the hand detection process. The hand likelihood image can contain unwanted static artefacts, due to skin coloured objects or shadows in the image. We can remove these artefacts by applying a background segmentation algorithm \citep{Zivkovic04}, which classifies pixels as foreground or background objects using an adaptive per pixel Gaussian mixture model. This segmentation process labels static objects as background by maintaining a history of pixel values over frames, so assumes a static camera. As a result, this assumption may not be ideal for applications in mobile robotics. Fortunately, the background segmentation is not essential and can be removed for mobile applications, with only a slight degradation in qualitative hand detection results. 

Immediately after initialisation, a mean-shift tracker \citep{Bradski98} is used to track the detected hands. Mean-shift locates the maxima of a likelihood function, in this case the re-projection likelihood obtained using the detected face, using discrete samples from the distribution. On each iteration, the original hand position is adjusted based on the mean-shift maxima.  As hands typically form larger blobs than wrists, the mean-shift tracker tends to remain centred on hands, and does not typically move along skin coloured forearms. When combined with the initialisation process, this allows for relatively robust hand tracking. If hands are lost (the average likelihood in the tracked hand area drops below a predefined threshold), the user simply re-initialises the hand tracker by returning their hands to the original initialisation area. 

\begin{figure}[!ht]
	\centering
	\subfloat[Likelihood map]{\includegraphics[width=0.3\textwidth]{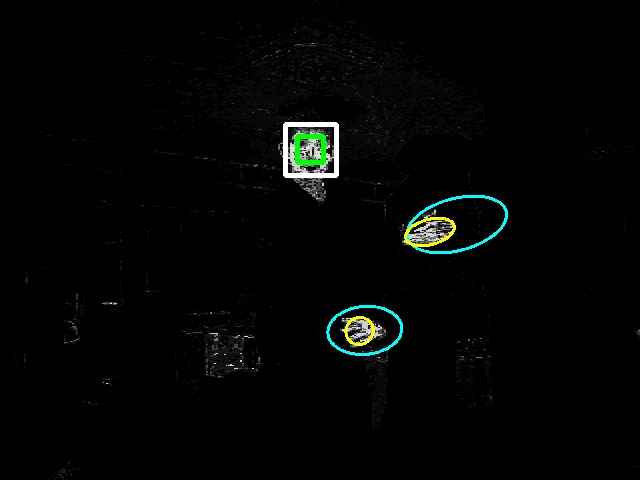}\label{fig:likely}}
	\subfloat[2D Pose estimate]{\includegraphics[width=0.3\textwidth]{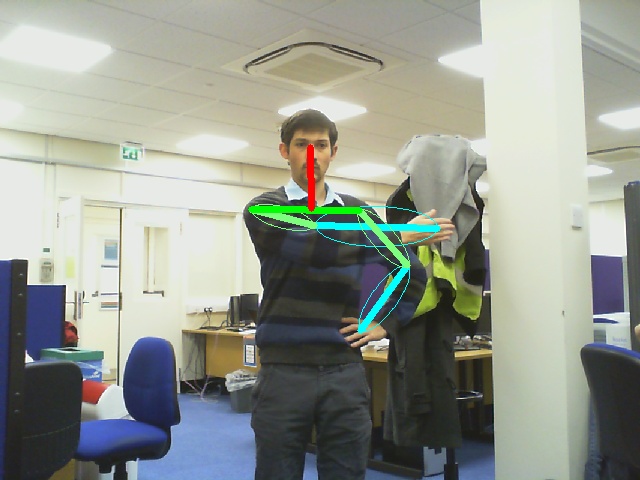}\label{fig:2Dpose}}
	\subfloat[3D Pose estimate]{\includegraphics[width=0.29545\textwidth]{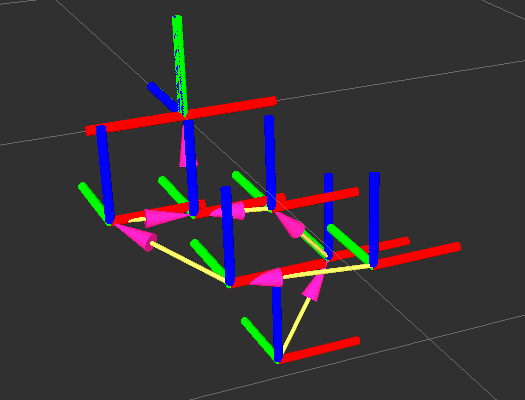}\label{fig:3Dpose}}
	\caption{An image patch (green square) centred in the detected face (white square) is used to build a likelihood map of skin coloured areas in the image (Figure \ref{fig:likely}). Detected hands are shown using yellow ellipses, with cyan ellipses showing the predicted hand locations. Once detected, the head and hand estimates are used to update the mixture Kalman filter and provide 2D (Figure \ref{fig:2Dpose}) and 3D (Figure \ref{fig:3Dpose}) pose estimates.}
	\label{bpHandTrack}
\end{figure}

The mean-shift tracker is unable to track rapidly moving objects particularly well, so is augmented through the use of a constant velocity Kalman filter tracker similar to that used for face tracking, which provides a predicted region of interest in which to search for a hand and improves the mean-shift tracking. Note that the predicted region could have been obtained by using the predicted body position in the mixture Kalman filter framework, but it turns out that the random walk motion model is not a very good predictor of hand positions, since it contains no velocity information. Figure \ref{bpHandTrack} illustrates the detection and tracking process. 

\begin{figure}[!ht]
	\centering
	\subfloat[]{\includegraphics[width=0.22\textwidth]{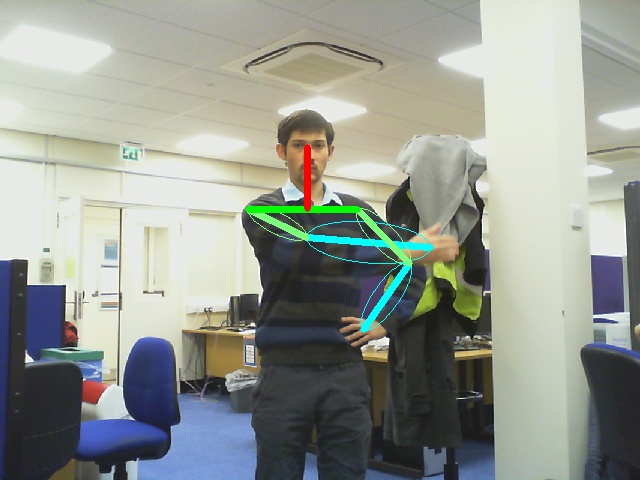}\label{mergea}}
	\subfloat[]{\includegraphics[width=0.22\textwidth]{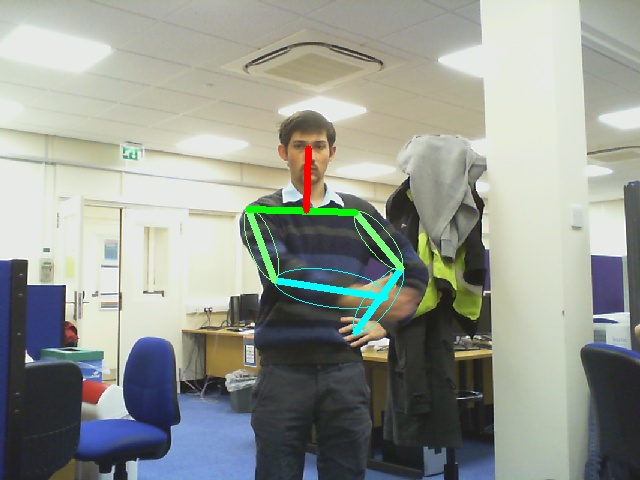}}
	\subfloat[]{\includegraphics[width=0.22\textwidth]{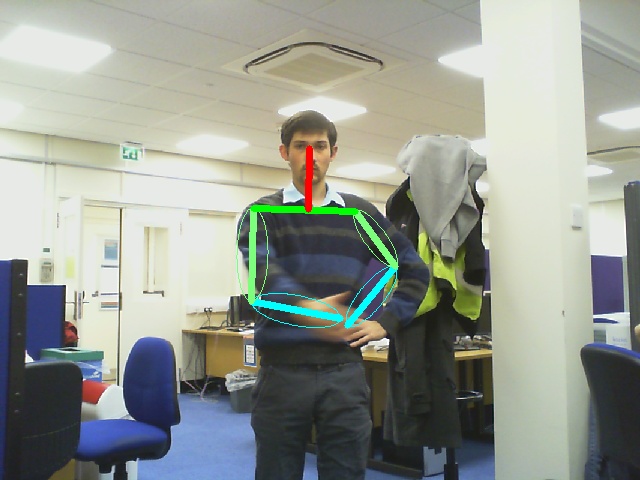}}\\
	\subfloat[]{\includegraphics[width=0.22\textwidth]{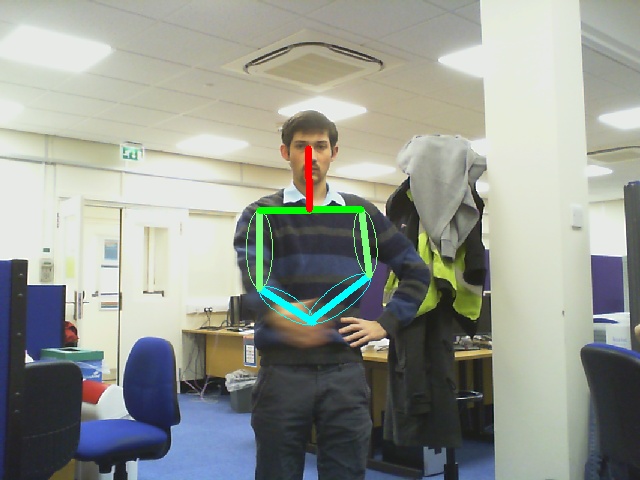}}
	\subfloat[]{\includegraphics[width=0.22\textwidth]{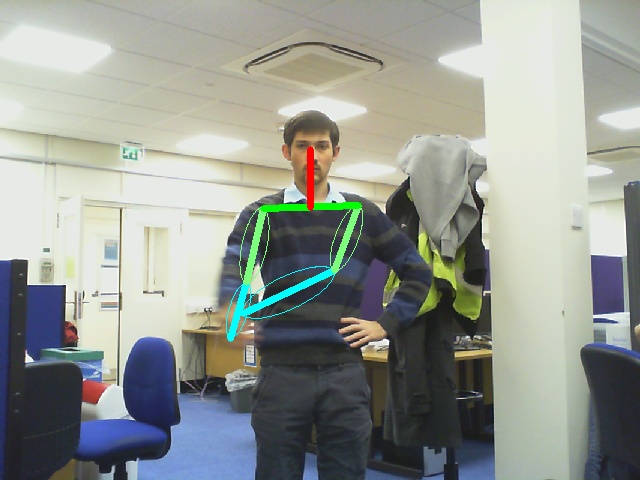}}
	\subfloat[]{\includegraphics[width=0.22\textwidth]{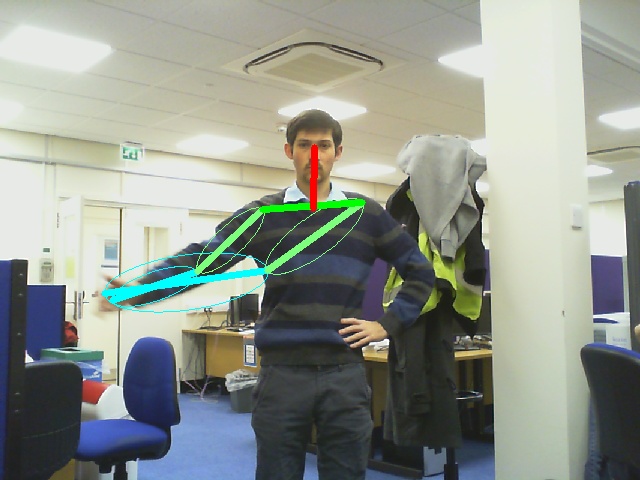}\label{mergeb}}\\
	\subfloat[]{\includegraphics[width=0.22\textwidth]{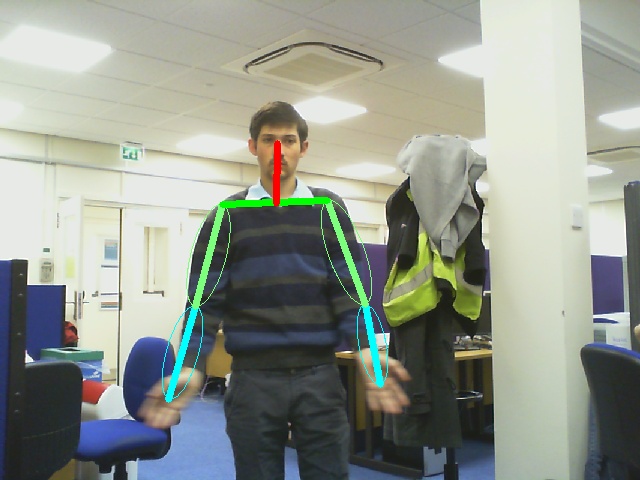}\label{tracka}}
	\subfloat[]{\includegraphics[width=0.22\textwidth]{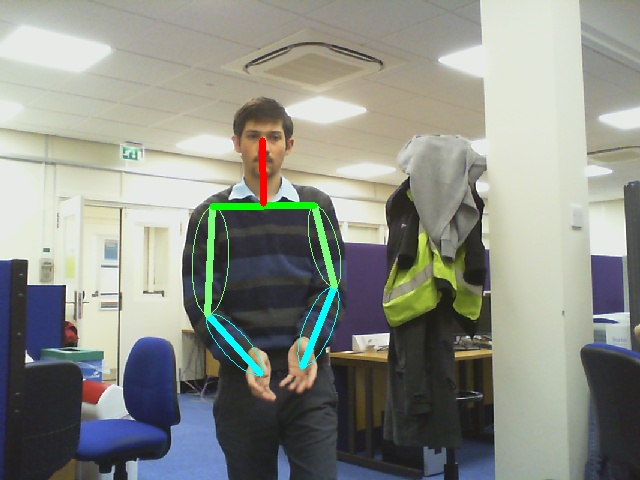}}
	\subfloat[]{\includegraphics[width=0.22\textwidth]{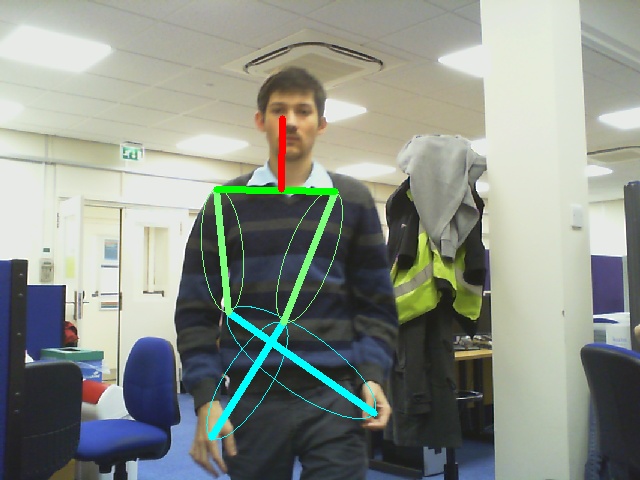}\label{trackb}}
	\caption{Similarities in hand appearance occasionally result in merging hands and tracking failures. Figures \ref{mergea} to \ref{mergeb} show a failure due to merging hands, while Figures \ref{tracka} to \ref{trackb} highlight a failure to track a clapping motion, resulting from the constant velocity motion model. A better hand detector or knowledge of forearm position could be used to remedy this.\label{handtrackerrors}}
\end{figure}

Unfortunately, the use of skin colour to detect hands leads to difficulties in discriminating between hands. This is alleviated somewhat by masking the image area predicted to contain the left hand when tracking the right hand, and vice versa with the left, but problems still occur when hands merge, or for clapping motions, where a constant velocity prediction causes hands to swap. Examples of these failures are shown in Figure \ref{handtrackerrors}. \citet{micilotta04} have proposed the use of a GMM trained using prior pose estimates to disambiguate left and right hands, but this simply tends to identify hands as left if they are found to the left side of the head (and vice versa to the right), sometimes incorrectly rejecting instances where hands cross the body.


\subsection{Edge-based hand association}

Errors resulting from incorrect hand association could be avoided by taking the orientation of the arms into account in the hand detection process. Unfortunately, it is quite difficult to detect arms, which can have highly variable appearances in images. However, once hands have been detected, we can assess the validity of a pose estimate using additional image features and use this to correct hand association errors. The MKF pose estimate contains the 2D position of each joint, and can be used to form a stick model similar to that drawn in Figure \ref{handtrackerrors}, with limbs described by a set of oriented edges. As a result, we propose that a natural measure of a pose estimate's likelihood is one that uses orientation information from edges detected in the image. 

\begin{figure}[ht]
	\centering
	\subfloat[Correct pose estimate]{\includegraphics[width=0.5\textwidth]{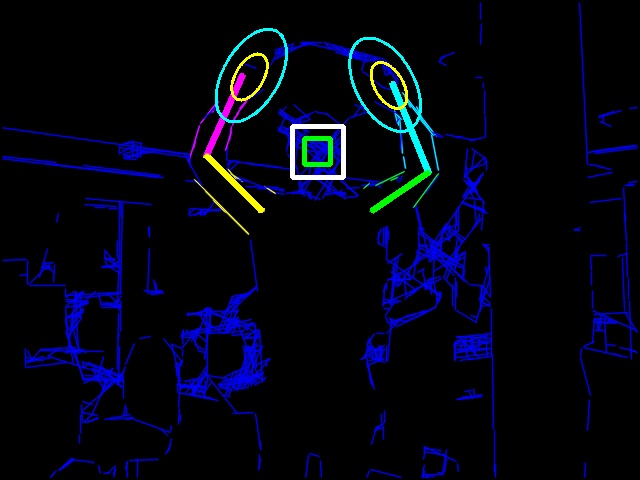}\label{fig:goodEdge}}
	\subfloat[Incorrect pose estimate]{\includegraphics[width=0.5\textwidth]{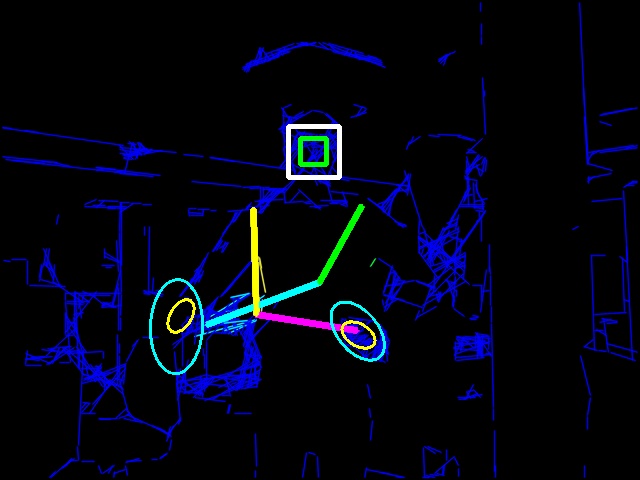}\label{fig:badEdge}}
	\caption{Detected image edges are used to determine the validity of a pose estimate. Edges providing no evidence are blue, while supporting edges are drawn in the same colour as the supported limb. In the case of valid pose estimates (Figure \ref{fig:goodEdge}), a number of edges with similar position and orientation to the estimated limb position tend to be observed. This typically fails to occur when a hand association error has occurred or if a pose estimate is incorrect (Figure \ref{fig:badEdge}).\label{edgeVoting}}
\end{figure}

Initially, an edge-based image representation is obtained using the Canny edge detector \citep{Canny86}. The probabilistic Hough line detector \citep{Matas2000} is then used to detect linear edge segments. The number of edge segments providing support for a pose estimate or limb position is then used to decide if the correct hand association has been made, or if the pose estimate is in error. We use a Gaussian kernel to determine edge support, with edges considered as evidence for a given limb if the likelihood 
\begin{equation}  
\mathcal{N}\left(\mathbf{x}_{\text{edge}}|\mathbf{x}_{\text{pose}}, \mathbf{\Sigma}\right) > \tau
\end{equation}
is greater than some threshold $\tau$. Here, $\mathbf{x}_{\text{edge}}$ is a vector of the edge orientation and the $x,y$ image position of a detected edge midpoint, while $\mathbf{x}_{\text{pose}}$ contains the position and orientation of the estimated limb. $\mathbf{\Sigma}$ is a diagonal covariance matrix, with variances selected empirically to allow feasible position and angle offsets. Figure \ref{edgeVoting} illustrates the voting process for a given pose estimate.

The proposed heuristic allows for data association errors in hand measurement to be corrected relatively quickly, but does not prevent these errors from occurring in the first place. Direct measurement of limb positions should eliminate hand association errors of this type completely.
 
\section{Combined detection and tracking results}
\label{Results}

\begin{figure}[!ht]
	\centering
		\includegraphics{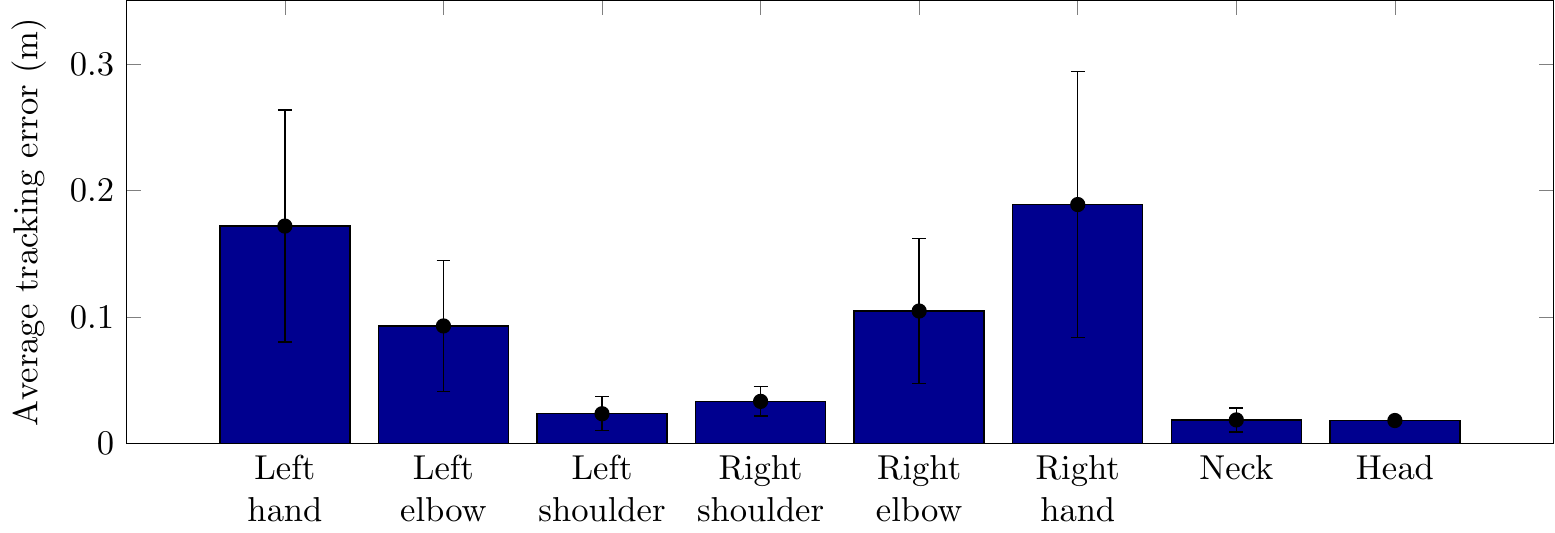}
	\caption[Average joint error in 3D]{Average 3D joint error over an image sequence containing a moving person.} \label{avetrackerror3D}
\end{figure}

Results obtained when the head and hand detectors of Section \ref{headhandmeasure} are used in conjunction with the mixture Kalman filter (deterministically selected tracks) are provided here. Figure \ref{avetrackerror3D} shows the mean error for each joint over a test sequence of more than 1000 images, when estimated 3D positions were compared with the skeleton output of a Kinect sensor, by aligning the head, neck and shoulders using fixed scale Procrustes analysis \citep{procrustes66}. This comparison is not ideal, as the Kinect is not perfectly accurate and often fails when hands cross over the body, but it does provide an indication that the 3D pose estimate is plausible.

\begin{figure}[!ht]
	\centering
	\includegraphics{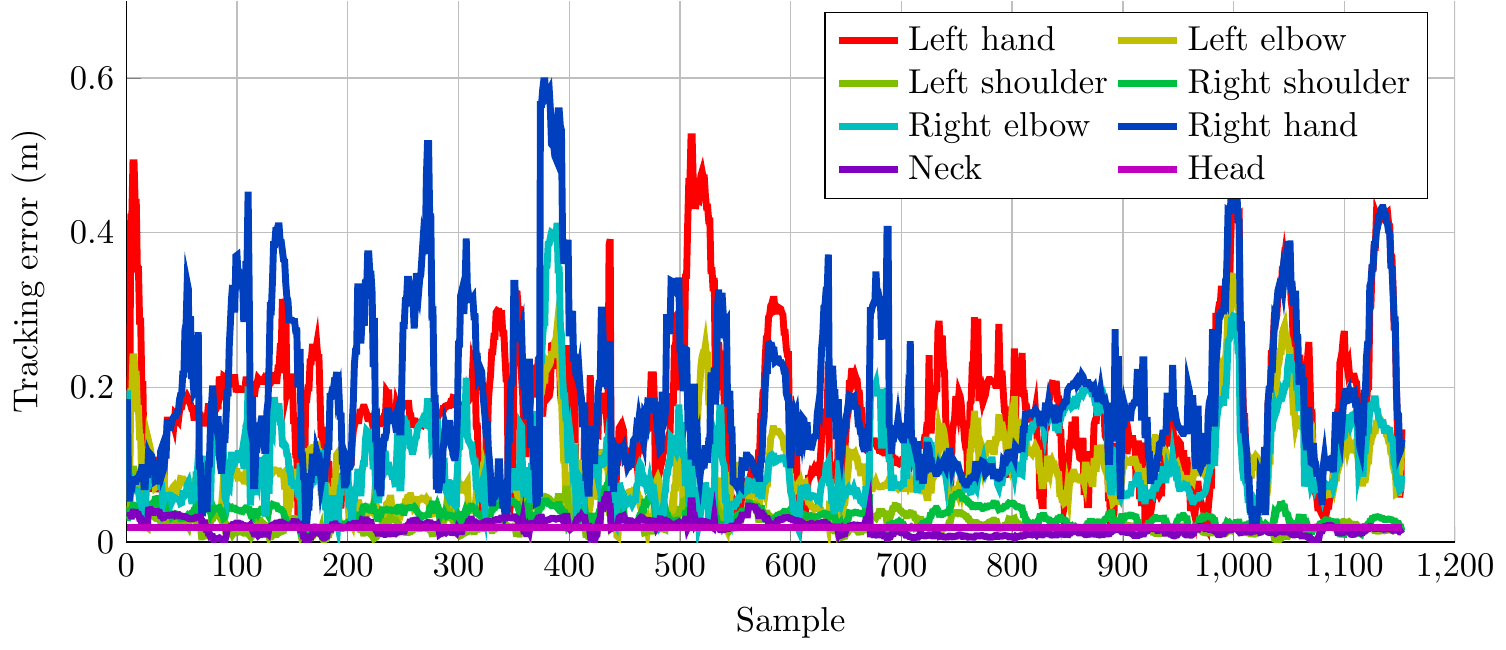}
	\caption[3D Tracking error for a moving upper body pose]{3D Tracking error for an image sequence of a moving upper body.} \label{trackerror3D}
\end{figure}

Figure \ref{trackerror3D} shows the position errors obtained for each joint over the entire test period, when compared with the positions obtained with the Kinect sensor. For much of the time, the error in hand position remains below 20 cm, with error spikes only occurring when the hands crossed the body or moved rapidly. A particularly encouraging result is that the average elbow errors remained quite low, even though no measurements of these joints were made at all. Qualitative results can be seen in the accompanying video, which also shows the edge-based error correction in operation.

A noticeable source of error involved uncommon poses that were not present in the prior training data. This should be remedied by additional training, but potentially at the expense of pose estimation accuracy in other poses.
\begin{figure}[!ht]
	\centering
	\includegraphics{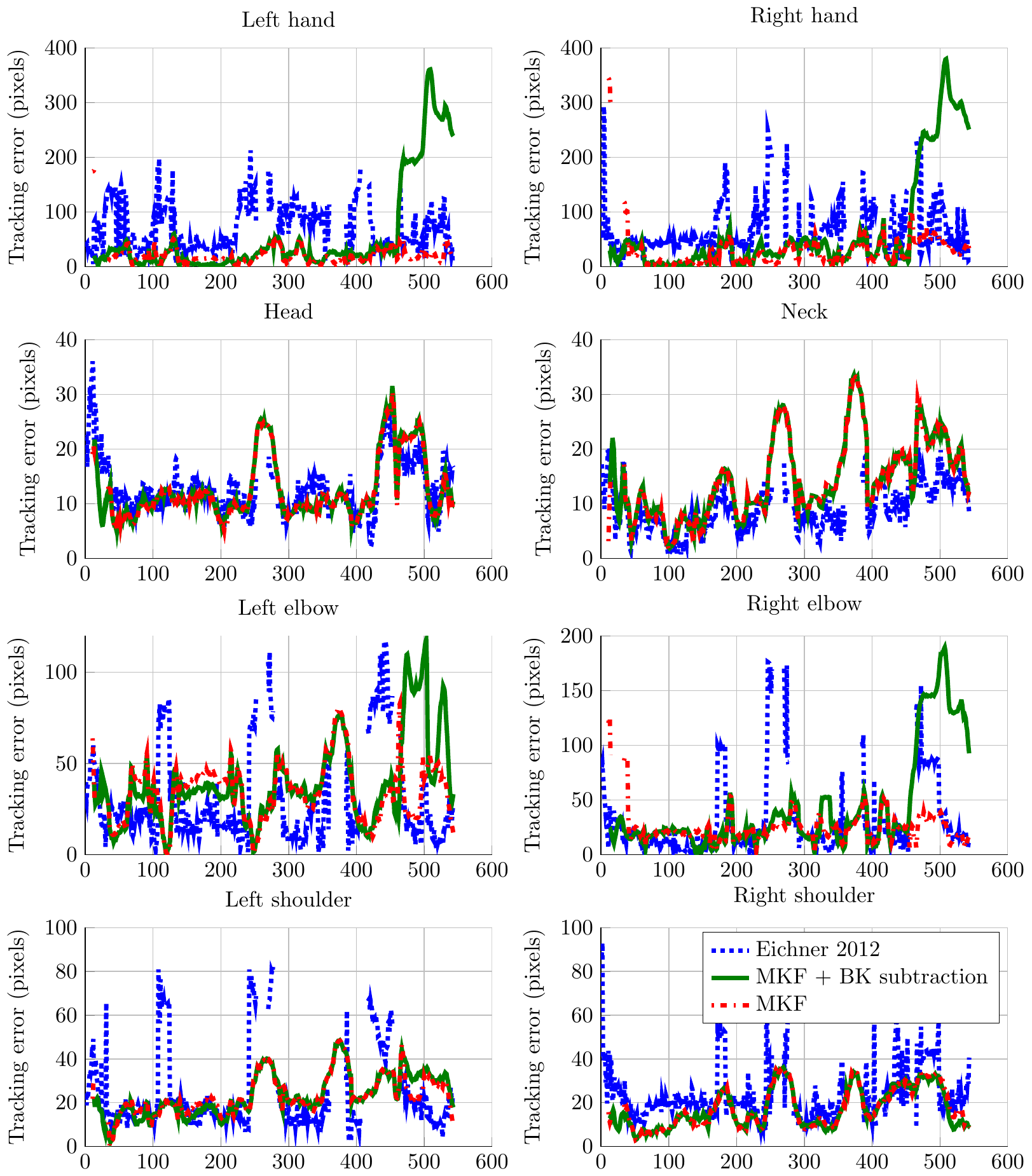}
	\caption[Tracking error in 2D]{2D pose errors over time obtained by applying the pose estimation of \citet{Eichner12} and our MKF approach, with and without background subtraction.} \label{eichTrack}
\end{figure}

Figure \ref{eichTrack} shows the 2D tracking errors obtained by applying the \citet{Eichner12} 2D pose estimation approach (the current state of the art in 2D pose estimation in unconstrained images) and our technique to a sequence of over 500 images, using Kinect joint tracks as ground truth. This sequence is particularly challenging for our approach as the study participant is wearing short sleeves, which could potentially result in hand detection failures due to skin coloured arm regions, and increases the risk of incorrect hand association.  The latter occurs towards the end of the tracking sequence, resulting in large hand and elbow tracking errors that failed to be corrected immediately by the edge-based pose correction. The figure also shows the results of the MKF pose estimation when background subtraction is not applied, which are very similar to those obtained when this is included. In fact, additional noise in the likelihood map used for hand detection prevented the hand association failure that occurred when background subtraction was applied, resulting in overall improved performance, although a number of spurious pose estimates were observed instead. 

It should be noted that the \citet{Eichner12} approach is at a disadvantage here as it does not incorporate temporal information, but it does perform far more processing, operating at approximately 0.5 frames a second. Our approach operates at just under 30 frames per second, with face detection the primary bottle neck.

\begin{figure}[ht]
	\centering
	\includegraphics{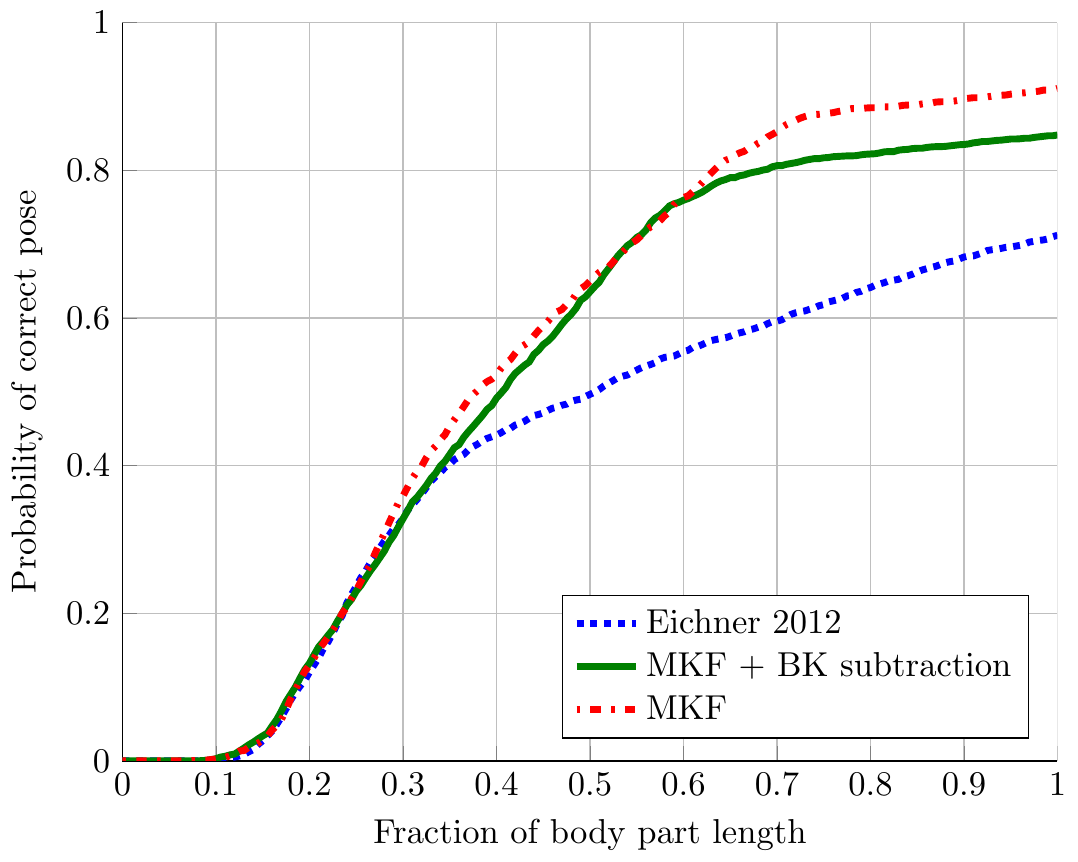}
	\caption{Probability of correct pose (PCP) curves for the MKF (with and without background subtraction) and \citet{Eichner12} approaches show significantly better performance when our approaches are used.}	\label{fig:pcpEich}
\end{figure}

In practise, the \citet{Eichner12} pose estimation performed well at upper arm detections, but typically failed at forearm detections, presumably due to the cluttered background used for experimentation. Figure \ref{fig:pcpEich} shows the PCP curves comparing the 2D pose estimation accuracy. Our approach shows significant improvement in detection rates for greater detection thresholds, while providing similar performance to \citet{Eichner12} over smaller thresholds. Once more, improved performance was seen when background subtraction was not applied, resulting from the absence of incorrect hand association in this test set, but a performance reduction is expected in cases where a number of skin-coloured objects are present in the image background. Qualitative results can be seen in the accompanying video, which provides a comparison with Kinect pose estimation and that of \citet{Eichner12}.

\section{Conclusions and future work}
\label{sec:concs}

This paper has provided results on upper body pose tracking using Kinect joint priors and simple hand and head measurements. Four tracking schemes have been considered and a mixture Kalman filter shown to provide effective upper body pose estimation. The use of the proposed upper body model allows reliable pose estimates to be obtained indirectly for a number of joints that are often difficult to detect using traditional object recognition strategies. The suggested model is designed with computational efficiency and analytical tractability in mind, yet still incorporates bio-mechanical properties of the upper body, typically only included using more complex body models.  

Comparisons with the current state of the art in 2D pose estimation \citep{Eichner12} have shown that our approach outperforms this significantly, both in terms of estimation performance and time complexity. Good 3D tracking results were also exhibited during experimentation.

A mechanism for correcting hand data association errors has been provided, but these errors will continue to occur without the inclusion of additional joint measurements. Improved hand association is required if multiple humans are to be tracked at once. While good results have been obtained for a constrained set of camera viewpoints, additional priors and improved measurements may be required to resolve pose ambiguities if 3D position is required over a larger range of viewpoints. Future work will involve the inclusion of a better mechanism for detecting hands and evaluating the effects of including additional training data collected from multiple persons. 

\bibliographystyle{abbrvnat}
\renewcommand{\bibname}{References} 
\bibliography{references} 

\appendix
\renewcommand*{\thesection}{\Alph{section}}
\section{Gaussian mixture models}
\label{GMMTraining}
Many tracking applications require a suitable probabilistic model of prior and likelihood distributions. Gaussian mixture models (GMMs) are a popular choice of model for probability distributions due to their ability to approximate a wide variety of complex distributions with a limited number of parameters. Only a brief overview  of GMMs is provided here, but readers are referred to \citet{Bishop06} for additional information. These models are particularly useful in acquiring an analytical approximation to a probability distribution when only discrete samples from the distribution are available. Formally, a Gaussian mixture model is defined as
\begin{equation}
p\left(\mathbf{x}_t\right) = \sum_{k=1}^{N_d} \pi_k p_k\left(\mathbf{x}_t\right), \label{prioreq}
\end{equation}
where
\begin{equation}
p_k\left(\mathbf{x}_t\right) = \frac{1}{\left(2\pi\right)^{d/2}|\mathbf{\Sigma}_k|^{1/2}}\text{exp} \left( -\frac{1}{2}  \left( \mathbf{x}_t - \boldsymbol\mu_k \right)^\text{T} \mathbf{\Sigma}_k^{-1}  \left( \mathbf{x}_t - \boldsymbol\mu_k \right) \right),
\end{equation}
with $N_d$ parameters $\boldsymbol\mu_k$, $\mathbf{\Sigma}_k$ and $\pi_k$. $d$ denotes the length of the state vector $\mathbf{x}_t$. $\mathbf{\Sigma}_k$ is symmetric and positive definite.

Training a GMM using discrete data is accomplished through expectation maximisation. Expectation maximisation is an iterative two step process obtaining the maximum likelihood estimation of parameters in a model. Assuming $N$ observations, start with an initial, random estimate of the model parameters and calculate the responsibility that the $k$-th Gaussian takes for explaining an observation $\mathbf{x}_i$,
\begin{equation}
\gamma_{ik} = \frac{\pi_k p_k\left(\mathbf{x}_i\right)}{\sum_{j=1}^{N_d} \pi_j p_j\left(\mathbf{x}_i\right)}.
\end{equation}
This is termed the expectation step. The maximisation stage occurs by applying analytic estimators to maximise the likelihood of the data. Parameter $\boldsymbol\mu_k$ is calculated as
\begin{equation}
\boldsymbol\mu_k = \frac{1}{N_k} \sum_{i=1}^N \gamma_{ik} \mathbf{x}_i;
\end{equation}
and $\mathbf{\Sigma}_k$ as
\begin{equation}
\mathbf{\Sigma}_k = \frac{1}{N_k} \sum_{i=1}^N \gamma_{ik}  \left(\mathbf{x}_i - \boldsymbol\mu_k\right) \left(\mathbf{x}_i - \boldsymbol\mu_k\right)^\text{T}.
\end{equation}
The effective number of points assigned to the $k$-th Gaussian in the mixture model is calculated as
\begin{equation}
N_k = \sum_{i=1}^N  \gamma_{ik}.
\end{equation}

\end{document}